\documentclass[a4paper,onecolumn,notitlepage,nofootinbib,longbibliography,superscriptaddress,floatfix]{revtex4-2}

\usepackage{xcolor}
\usepackage[colorlinks=true,citecolor=teal,linkcolor=purple,urlcolor=purple]{hyperref}
\usepackage{url}            
\usepackage{booktabs}       
\usepackage{amsfonts}       
\usepackage{nicefrac}       
\usepackage{microtype}      
\usepackage{graphicx}
\usepackage{algorithmic}
\usepackage{algorithm}
\usepackage{amsmath}
\usepackage{amsthm}
\usepackage{amssymb}
\usepackage{wrapfig}
\usepackage{tabularx}
\usepackage{enumitem}
\usepackage{lipsum}
\usepackage{soul}
\usepackage{multirow}
\usepackage{parskip}
\usepackage{caption}
\usepackage{mathtools}

\newtheorem{theo}{Theorem}[section]
\newtheorem{prop}[theo]{Proposition}
\newtheorem{remark}[theo]{Remark}

\newcommand{\fu}{Dahlem Center for Complex Quantum Systems, Freie Universit\"{a}t Berlin, 14195 Berlin, Germany}
\newcommand{\pdg}{formerly at Porsche Digital GmbH, 71636 Ludwigsburg, Germany}

\begin{document}

\title{Using deep learning to construct stochastic local search SAT solvers with performance bounds}
\date{\today}

\author{Maximilian J. Kramer}
\email{m.kramer@fu-berlin.de}
\affiliation{\fu}
\affiliation{\pdg}

\author{Paul Boes}
\affiliation{\fu}
\affiliation{\pdg}

\author{Jens Eisert}
\affiliation{\fu}

\begin{abstract}
The Boolean Satisfiability problem (SAT), as the prototypical $\mathsf{NP}$-complete problem, is crucial in both theoretical computer science and practical applications. 
To address this problem, stochastic local search (SLS) algorithms, which iteratively and randomly update candidate assignments, present an important and theoretically well-studied class of solvers. Recent theoretical advancements have identified conditions under which SLS solvers efficiently solve SAT instances, provided they have access to suitable ``oracles'', i.e., instance-specific distribution samples. We propose leveraging machine learning models, particularly graph neural networks (GNN), as oracles to enhance the performance of SLS solvers. Our approach, evaluated on random and pseudo-industrial SAT instances, demonstrates a significant performance improvement regarding step counts and solved instances. Our work bridges theoretical results and practical applications, highlighting the potential of purpose-trained SAT solvers with performance guarantees.
\end{abstract}
\maketitle

\section{Introduction}

The application of deep learning to constraint optimization and constraint satisfaction problems has received
considerable attention in recent years, with promising results. This includes the application of deep learning to solve the
\emph{Boolean satisfiability problem} (SAT), as explored in various studies.
This includes the application of deep learning to solve the \emph{Boolean satisfiability problem} (SAT), with approaches ranging from survey works~\cite{guo2022SAT_ML}
and end-to-end neural solvers~\cite{buenz_lamm_2017graph, selsam2019learning, Ozolins_2022} to neural heuristics for guiding existing solvers~\cite{selsam2019guiding, jaszczur2020neural, han2020enhancing}
and learned components for local search~\cite{yolcu_2019, Zhang_2020} and CDCL-based solvers~\cite{wang2024}. We discuss the most relevant of these works in detail in Sec.~\ref{sec:related_work}.
SAT has famously been identified as 
the prototypical $\mathsf{NP}$-complete~ \cite{Cook1971, Levin1973} constraint satisfaction problem
in the Cook–Levin theorem. The $\mathsf{NP}$-completeness
of the problem does not imply that, for many instances, good solutions can be found, and substantial effort has been directed to identifying good solvers that work well in practice.
One practical motivation for the integration of \emph{machine learning} (ML) into solvers for $\mathsf{NP}$-hard optimization and decision problems is that ML models can learn structure that is present in the data for a given application, in which instances of the problem might differ in the details but still be very similar in terms of their structure. 

\begin{figure}
  \centering
  \includegraphics[width=0.8\textwidth]{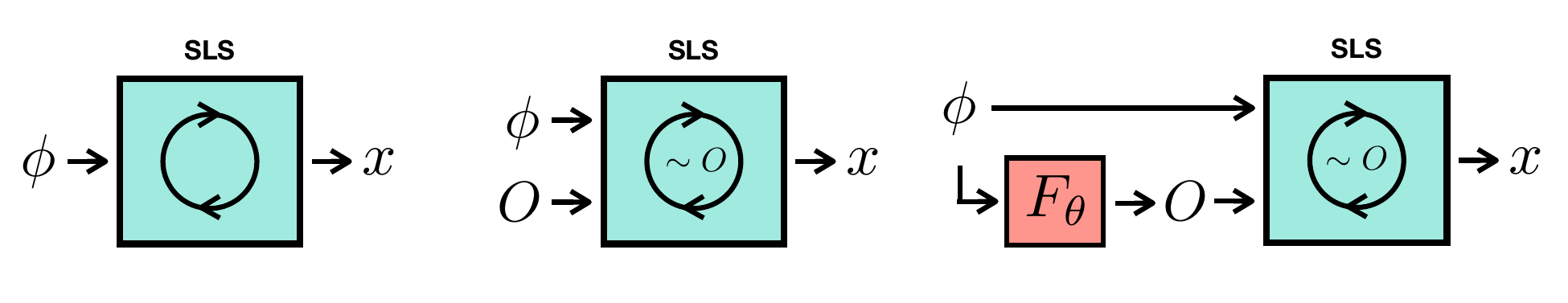}
  \caption{Illustration of the general idea of this work. \emph{Left:} A simple SLS solver finds a solution to a SAT instance by repeatedly and randomly updating a small subset of the variables. \emph{Middle:} An oracle-based SLS solver uses samples from an oracle $O$ provided as input to update the variables at each iteration. \emph{Right:} We use a deep learning model to train an \emph{oracle factory} $F_\theta$ on distributions of instances. This oracle factory
  maps an incoming instance to an oracle, which is then fed into an oracle-based SLS solver. This approach is motivated by results that provide sufficient conditions for an oracle-based SLS solver to find a solution efficiently based on the properties of the oracle.
  }
  \label{fig:general_idea}
\end{figure}

There are many ways of bringing ML into solvers. The main idea of this work is to use a deep learning model, in our case a \emph{graph neural network} (GNN), as an \emph{oracle factory} that, given a SAT instance, outputs a \emph{sampling oracle}, or simply puts an instance-specific distribution that is fed to a \emph{stochastic local search} (SLS) solver, a particular type of random walk algorithm. During the search process, the SLS solver samples from this distribution in an offline fashion. This approach stands in contrast to other works that use the ML model either online while running the algorithm, which renders the approach often inefficient, or 
only to generate an initial candidate solution fed to an off-the-shelf solver. In Sec.~\ref{sec:methods}, we provide details of our approach, while Fig.~\ref{fig:general_idea} provides a rough sketch.

The main source of inspiration for the approach taken in this work is a series of breakthrough results from theoretical computer science (see Refs.~\cite{Moser2008, Moser2009, harris2017algorithmic, achlioptas2020lovasz}) that imply sufficient conditions for an SLS solver, in case it has access to a suitable oracle, to efficiently find a solution to a SAT instance. By pursuing an approach that originates from theoretical considerations, this approach also considers SLS solvers' well-studied abilities and limitations (see Refs.~\cite{Schoening1999, schöning2013satisfiability}) and addresses them appropriately.

\subsection{Contributions}

This work makes a number of original contributions:

\begin{itemize}[noitemsep]
    \item We establish the notion of oracle-based SLS solvers by explicitly constructing and training two oracle-based SLS solvers using an 
    \emph{interaction network}~\cite{battaglia2016interaction}. These algorithms have the advantage of an offline use of the ML model while going beyond initialization, 
    e.g., warm-starts.
    \item We establish the promise of application-specific solvers with performance guarantees by connecting ``hands-on'' ML research with theoretical results from computer science. 
    \item We elaborate on the insight
    that an oracle does not need to be perfect, or close to being perfect, to increase the performance of an SLS solver significantly. On the one hand, we motivate the search for ``sufficiently good'' oracles by theoretical results from computer science. This includes introducing a new, theoretically motivated, loss function to study random SAT instances, which we call the Lov\'{a}sz Local Lemma loss, that rewards an oracle's exploitation of the local structure of SAT instances. We are not aware of any work in the field whose loss is motivated by known bounds on the resulting solver performance. On the other hand, we discuss a family of SAT instances that are provably hard for SLS solvers and develop conditions on ``sufficiently good'' oracles such that our oracle approach finds the solution in polynomial and even linear time.
    \item We empirically investigate the ability of these solvers to solve random $3$-SAT at varying levels of difficulty (as measured by the common $\alpha$-ratio between clauses and variables). Our experiments show significant boosts in performance, with the ML-based solvers 
    solving, on average, $14\%$ more instances and doing so in $33\%$ fewer steps, with around $6.7\times$ improvement in the median number of steps. We show that the ML-based solver has a significantly higher algorithmic barrier than the uniform version.
    \item We show that, in our experiments, providing continuous access to an oracle produces significantly better results (both in the number of steps needed and the number of instances solved). In contrast, an algorithm that only uses the oracle to initialize a candidate but keeps the uniform update rule only solves instances in fewer steps and cannot solve more instances.
    \item Furthermore, we study the pseudo-industrial benchmark sets with instances of varying difficulty introduced in Ref.~\cite{li2023g4satbench}. Here, we show that the learned oracle also leads to a significantly stronger solver in terms of instances solved and the number of steps taken.
\end{itemize}

\subsection{Related work}
\label{sec:related_work}
\paragraph{Deep learning based SAT solving:}
Ref.~\cite{guo2022SAT_ML} provides a recent review paper on machine learning-based SAT solvers. Focusing here on the literature on SLS solvers, in Ref.~\cite{Zhang_2020}, the authors train a Graph Neural Network to generate an initial candidate for various SLS solvers. However, they do not use this GNN as an oracle while running these solvers. Ref.~\cite{yolcu_2019} uses a GNN to learn a variable selection heuristic in a WalkSAT-type SLS solver. However, their model is based on a learned policy and hence does not act as an oracle in the sense of this work. 
Despite the literature on SLS solvers, it should be noted that Ref.~\cite{wang2024} has with our work in common that a GNN-prediction is queried once and then used in an offline-fashion while running the solver. However, the idea is applied only to CDCL solvers.
Recently, Ref.~\cite{Skenderi2026} proposed hard CSP benchmarks from a statistical-physics perspective and observed that classical heuristics still outperform GNN-based solvers on such instances. Our work is complementary: rather than claiming GNN superiority over classical solvers, we study how theory-guided oracles can improve a given SLS solver relative to its uniform baseline.
Moreover, the cross-entropy loss with respect to a Gibbs distribution that we use in this work is also used in Ref.~\cite{Deepmind2020} in the context of \emph{mixed integer program} solving. 

\paragraph{SLS algorithms with solution guarantees:}
As described above, one core contribution of this work is to choose a solver and loss function for training such that we are guaranteed by theoretical results that a smaller loss will lead to better performance. The body of work we have in mind here is that around the seminal \emph{Lov\'{a}sz Local Lemma} (LLL)~\cite{Erdos_Lovasz1975}. In particular, Refs.~\cite{Moser2008, Moser2009} have proven those results that underlie our Proposition~\ref{prop:moser}. Ref.~\cite{harvey2015algorithmic} explicitly introduces the notion of an oracle-based SLS algorithm, even though their \emph{re}sampling oracles are rather special cases of the oracles we consider here. Ref.~\cite{harris2017algorithmic} proves the results stated below as Proposition~\ref{prop:harris} and that we consider as crucial for motivating our work in settings where an oracle does not satisfy the conditions of the LLL.

The \emph{WalkSAT algorithm}~\cite{selman1996local} 
has been an object of intense theoretical studies, with worst-case bounds on the runtime for 3-SAT and 2-SAT derived in Refs.~\cite{Schoening1999, Papadimitrou}, respectively.
Moreover, the existence of an algorithmic barrier for WalkSAT has been studied, e.g., in Refs.~\cite{semerjian2003study, Alekhnovich_2007, Coja_Oghlan_2014, Coja_Oghlan_2017}. However, it should be noted that none of these results uses the notion of an oracle. For this particular algorithm, SAT formulas that are hard to solve are established, with an explicit result and analysis given in Ref.~\cite{schöning2013satisfiability}. Let us emphasize the oracle-like working of the \emph{ProbSAT} algorithm from Ref.~\cite{probsat} that has served 
as an inspiration for Algorithm~\ref{alg:oracle_based_walksat}.

\section{Methods}
\label{sec:methods}
\subsection{SAT}
The \emph{SAT decision problem} is to decide whether a given formula $\phi$, that is, a formula consisting of $n$ Boolean variables connected by the Boolean operators (conjunction $\wedge$, disjunction $\vee$ and negation $\neg$), admits a truth value assignment $x \in \{0,1\}^n$ to the $n$ variables such that the formula, as a whole, evaluates to true under this assignment. For every formula, there exists an equivalent formula that is in \emph{conjunctive normal form} (CNF), which is a conjunction of clauses,
\begin{align}
    \phi = (c_1 \wedge c_2 \wedge \dots \wedge c_m), \text{ with } c_j = (l^{(j)}_1 \vee l^{(j)}_2 \vee \dots \vee l^{(j)}_k),
\end{align}
where each clause $c_j$ is a disjunction of literals (i.e., variables or their negations). SAT is $\mathsf{NP}$-complete in worst-case complexity by virtue of the Cook–Levin theorem (see Refs.~\cite{Cook1971, Levin1973}). At the same time, it has many applications in industry. To name a few, it can be used to verify hardware and software, to plan and schedule, and to solve problems in cryptography. 

Notationwise, in the following we write $[n] \coloneq \{1, \dots, n\}$, $|\phi|$ for the number of clauses present in $\phi$, $c \in \phi$ to denote that the clause $c$ is present in a CNF-formula $\phi$, and $V(c) \subseteq [n]$ to denote the set of variables present in $c$. We further split $V(c)$ into disjoint subsets $V^-(c)$ and $V^+(c)$, depending on whether the variables appear in $c$ with or without negation, respectively.\footnote{Note that we can assume, w.l.o.g., that no variable appears twice in any clause, since we can in this case generate an equivalent formula $\phi'$ that satisfies this assumption, in linear time.} For any assignment $x$ and any subset $V\subseteq [n]$ of variables, we write $x|_V = (x_v)_{v \in V}$ to denote the truncated assignment on $V$, $c(x) \in \{0,1\}$ to indicate whether a clause is violated or not, and $\phi(x) = \sum_j c_j(x)$ to denote the number of violated clauses. Finally, we write $X_n \coloneq \{0,1\}^n$ and, for an instance $\phi$, we denote as $\Pi(\phi) \subset X_n$ the set of assignments with the minimal number of violated constraints across $X_n$. An instance is \emph{satisfiable} if there exists an $x$ such that $\phi(x) = 0$ and \emph{unsatisfiable} otherwise.

\subsection{Oracle-based SLS}
\label{sec:oracle_based_solvers}
Many solvers exist to address the general SAT problem. One class of them are \emph{stochastic local search} (SLS) algorithms. In their simplest form, they work as follows:
\begin{enumerate}[noitemsep]
    \item Given a SAT instance $\phi$ in \emph{conjunctive normal form} (CNF), generate an initial candidate $x$ via an \texttt{initialize} sub-routine.
    \item If $x$ violates any clause, choose one violated clause $c$ at random and update $x$ on the variables that appear in $c$ via a sub-routine \texttt{update}, keeping the remainder of $x$ unchanged.
    \item Repeat step 2 until no clause is violated or a stopping criterion is met. Output $x$.
\end{enumerate}

Hence, specifying a simple SLS solver amounts to specifying the \texttt{initialize} and \texttt{update} sub-routines. Possibly the two simplest non-trivial SLS algorithms are the \emph{Moser-Tardos} (MT) algorithm and the simple \emph{WalkSAT} algorithm. The MT algorithm initializes an assignment randomly and updates an assignment $x$ by randomly sampling a new assignment $x'$ and setting $x$ to $x'$ on all variables in the clause $c$. WalkSAT~\footnote{We consider here the variant of WalkSAT, which coincides in its uniform version with Schöning's algorithm from Ref.~\cite{Schoening1999}. The heuristic of flipping with a certain probability the variable which minimizes the number of unsatisfied clauses is not used.} also initializes an assignment randomly; however, to update, it randomly chooses one of the variables that appear in $c$ and flips the assignment of that variable deterministically.

Both of the above algorithms heavily use samples drawn uniformly at random. However, it seems intuitively clear that they would perform better if, instead, they sampled from distributions that are fine-tuned to the instance $\phi$ and the current state $x$ of the solver. To formalize this, we introduce the notion of a (sampling) oracle. A \emph{sampling oracle for instance $\phi$} is a random variable $O$ over the sample space $\{0,1\}^n$. Oracle-based SLS solvers are then SLS solvers that accept an oracle as part of the input and whose sub-routines utilize either samples $x \sim O$ from $O$ or the values of the latter's probability measure $P_O$ in its definition. 
\begin{figure}[tb]
\begin{minipage}[t]{0.48\linewidth}
\centering
\begin{algorithm}[H]
\caption{Oracle-based Moser-Tardos algorithm}
\label{alg:oracle_based_moser}
\textbf{Input}: $\phi, O$\\
\textbf{Output}:  $x$
\begin{algorithmic}[1]
    \STATE $x \sim O$
    \WHILE{$\exists c \in \phi$: $c(x) = 1$}
        \STATE $x' \sim O$
        \STATE pick a violated clause $c \in \phi$ at random
        \STATE $x|_{V(c)} \leftarrow x'|_{V(c)}$ 
    \ENDWHILE
\end{algorithmic}
\end{algorithm}
\end{minipage}
\hfill
\begin{minipage}[t]{0.48\linewidth}
\centering
\begin{algorithm}[H]
\caption{Oracle-based WalkSAT algorithm}
\label{alg:oracle_based_walksat}
\textbf{Input}: $\phi, O$\\
\textbf{Output}:  $x$
\begin{algorithmic}[1]
    \STATE $x \sim O$
    \WHILE{$\exists c \in \phi$: $c(x) = 1$}
        \STATE pick a violated clause $c \in \phi$ at random
        \STATE pick variable $v$ from $c$ with probability $\frac{P_{O}(\neg x_v)}{\sum_{v' \in c}P_{O}(\neg x_{v'})}$
        \STATE $x_v \leftarrow \neg x_v$
    \ENDWHILE
\end{algorithmic}
\end{algorithm}
\end{minipage}
\end{figure}

We define the \emph{oracle-based MT} and the \emph{oracle-based WalkSAT} algorithms in Algorithm~\ref{alg:oracle_based_moser} and Algorithm~\ref{alg:oracle_based_walksat}, respectively. Both of these algorithms generalize their non-oracle-based counterpart since the former coincides with the latter when using the uniform oracle. The choice of sampling probabilities in the update set of Algorithm~\ref{alg:oracle_based_walksat} are inspired by the ProbSAT algorithm in Ref.~\cite{probsat} as they put more weight on choosing a variable $v$, the higher the likelihood under $P_O$ of sampling a state with a flipped value assignment. Here and in the remainder, for any set of variables $V$ and any $x_V \in \{0,1\}^{|V|}$, we write 
\begin{align}
P_O(x_V) = \sum_{x \in \{0,1\}^n: x|_V = x_V} P_O(x).
\end{align}
\subsection{Theoretical motivation to search for ``sufficiently good'' oracles}
\label{subsec:performance_guarantees}
It is clear that an oracle can be more or less suited to an instance, depending
a lot on the studied instance and the used solver. For example, for a given instance $\phi$, a ``perfect'' oracle, i.e., one that samples an element from $\Pi(\phi)$ with unit probability, leads to convergence of both of the above algorithms in a single step. Of course, perfect oracles are hard to come by since they imply solving the problem we address. As we will see below, an oracle does not need to be ``perfect'' to yield efficient solvers. We introduce the concept of ``sufficiently good'' oracles, motivated by two theoretical perspectives. These perspectives offer distinct criteria for defining what qualifies as a ``sufficiently good'' oracle.
\paragraph{MT algorithm:}
Fortunately, over the past 15 years, researchers from theoretical computer science have proven that the MT algorithm with access to a ``sufficiently good'' oracle is guaranteed to find a solution efficiently. We need to introduce some additional concepts and notation to formulate the existing performance guarantees for the MT algorithm. Let $O$ be an oracle, $\phi$ be a given Boolean formula in CNF with $n$ variables and $m$ clauses and let 
\begin{align}
    P_O(J|\phi) = \sum_{x \in X_n: c_{j}(x) = 1 \quad \forall j \in J} P_O(x)
\end{align}
be the probability that a sample $x \sim O$ violates the clauses $J \subseteq [m]$. We define the \emph{dependency graph} $\mathcal{D}_O(\phi)$ induced by this distribution as the graph that has $m$ nodes and with two nodes $j, j'$ being connected by an edge whenever they are not statistically independent, i.e.,
\begin{align}
    P_O(\{j,j'\}|\phi) \neq P_O(j|\phi) \cdot P_O(j'|\phi).
\end{align}
Given this graph, let $\Gamma(j)$ denote the exclusive neighborhood of node $j$ in this graph and $\Gamma^+(j) = \Gamma(j) \cup \{j\}$ as the inclusive neighborhood. We then have the following proposition, which is implied by the main result of Ref.~\cite{Moser2009}: 
\smallskip

\begin{prop}
\label{prop:moser}
Given a formula $\phi$ in CNF, if there exists a map $\mu: [m] \to [0,\infty)$ such that, for all $j \in [m]$, 
\begin{align}
\label{eq:LL_condition}
    P_O(j|\phi) \cdot \Pi_{j' \in \Gamma^+(j)} (1 + \mu(j')) \leq \mu(j), 
\end{align}
then $\phi$ is satisfiable and the MT algorithm with access to $O$ will find a solution to $\phi$ in an expected $\sum_j \mu(j)$ number of steps.
\end{prop}
In words, Proposition~\ref{prop:moser} states that if the probability of violating each clause under the oracle is sufficiently small relative to the local structure of the instance (as captured by the dependency graph), then the MT algorithm is guaranteed to find a solution efficiently.

The ``perfect'' oracle clearly satisfies this proposition, as $P_O(j|\phi) = 0$ for all $j$, so that we can set $\mu(j) = 0$ for all $j$. Importantly, however, imperfect oracles can still satisfy this condition. For instance, consider a $k$-SAT instance $\phi$ in which each clause carries at most $k$ literals and each variable appears in at most $\frac{2^k}{(k+1)e}$ clauses. Then, by setting $\mu(j) = e/2^k$ for all $j$ and using the fact that $(1 + 1/r)^r \leq e$ for all $r > 0$, we find that the uniform oracle over $X_n$ satisfies~(\ref{eq:LL_condition}) and hence the MT algorithm would find a solution to $\phi$ despite it being not tuned to $\phi$ at all. 
As such, Proposition~\ref{prop:moser} strongly motivates the search for oracles that satisfy~(\ref{eq:LL_condition}). Moreover, more recent results further motivate the construction of oracles that satisfy~(\ref{eq:LL_condition}) only approximately or for unsatisfiable instances. In particular, results in Ref.~\cite{harris2017algorithmic} imply the following:
\smallskip 

\begin{prop}
\label{prop:harris}
Given a formula $\phi$ in CNF and a map $\mu: [m] \to [0,\infty)$, there exists a simple extension to the MT algorithm that takes as input $\phi, \mu$ and an oracle $O$, that runs an expected $\sum_j \mu(j)$ number of steps and whose output $x$, for every $j \in [m]$, violates clause $c_j$ with probability
\begin{align}
    \mathrm{Pr}(c_j(x) = 1) \leq \max(0, \epsilon_{O,\mu}(j)),
\end{align}
where $\epsilon_{O,\mu}(j) = P_O(j|\phi) \cdot \Pi_{j' \in \Gamma^+(j)} (1 + \mu(j')) - \mu(j)$.
\end{prop}
Informally, the quantities $\epsilon_{\mathcal{O},\mu}(j)$ measure how far the oracle is from satisfying the sufficient condition of Proposition~\ref{prop:moser} on each clause. Our LLL loss given by Eq.~\eqref{eq:LLL_loss} directly penalizes these terms.

This proposition implies that it is in our interest to construct oracles that minimize the set of $\epsilon_{O,\mu}(j)$, even if we are not able to satisfy condition~(\ref{eq:LL_condition}) or work on satisfiable instances, such as in the case of the MaxSAT optimization problem, which is concerned with solvers that return elements in $\Pi(\phi)$ also for unsatisfiable instances.\smallskip

\begin{remark}
We emphasize that Propositions~\ref{prop:moser} and~\ref{prop:harris} provide guarantees for oracles that satisfy the stated conditions exactly. In practice, the oracle produced by our trained model approximates these conditions, with the LLL loss in Eq.~\eqref{eq:LLL_loss} directly penalizing the slack terms $\epsilon_{\mathcal{O},\mu}(j)$ from Proposition~\ref{prop:harris}. A smaller loss therefore corresponds to a closer approximation of the sufficient conditions and, by the proposition, to stronger performance guarantees for the resulting solver. The ablation study in Appendix~\ref{app:random3SAT_results} confirms that the LLL loss is the dominant contributor to the observed solver improvement for the random 3-SAT benchmark.
\end{remark}

\paragraph{WalkSAT:}
\label{subsec:motivating_example}
While we are not aware of general performance bounds for the WalkSAT algorithm analogous to the above propositions that use an oracle, researchers have investigated the existence of an algorithmic barrier for the uniform WalkSAT on random $k$-SAT. As usual, let $\alpha$ be the clauses to variables ratio. In particular, it has been shown in Ref.~\cite{Coja_Oghlan_2017} that WalkSAT is ineffective with high probability if $\alpha > c 2^k \ln(k)^2/k$ where $c > 0$ is an absolute constant. On the other hand, it has been proven in Ref.~\cite{Coja_Oghlan_2014} that WalkSAT finds satisfying assignments in linear time with high probability if $\alpha < c'2^k/k$ for another constant $c' > 0$. Moreover, arguments from physics in Ref.~\cite{semerjian2003study} suggest that WalkSAT is only effective up to $\alpha = (1 + o_k(1)) 2^k / k$ and not beyond. Ref.~\cite{Alekhnovich_2007} provides positive results for $k=3$ by proving linear upper bounds on the running time of this algorithm for $\alpha < 1.63$. However, experiments indicate an algorithmic barrier at around $\alpha = 2.6$, as reported in Ref.~\cite{Alekhnovich_2007} from a personal communication.

To establish the idea of oracles for WalkSAT, we provide an explicit example of a family of SAT instances that are provably hard for the uniform WalkSAT. However, we show that the oracle-version of WalkSAT converges in linear time to a solution if the algorithm has access to a suitable non-uniform oracle.
SAT formulas are well established to exist, which are hard for local search algorithms; for instance, Ref.~\cite{schöning2013satisfiability} introduces such an example 
\begin{align}
\label{eq:hard_formula}
    \phi_{\text{hard}}(n) = & c_{1} \wedge c_{2} \wedge \bigwedge_{\substack{(i,j,k)\in [n]^3:\\ j\neq i , k\neq i, j>k} } d_{(i,j,k)}
\end{align}
in Chapter 5.7, where $c_{1} = (\bar{x}_{1} \vee \bar{x}_{2} \vee \bar{x}_{3}), c_{2} = (\bar{x}_{4} \vee \bar{x}_{5} \vee \bar{x}_{6})$ and $d_{(i,j,k)} = (\bar{x}_{i} \vee x_{j} \vee x_{k})$. 
For $n$ variables, there are exactly $l = \binom{n}{3} \cdot 3$ clauses of type $d_{(i,j,k)}$. The intuition behind the clauses $c_1$ and $c_2$ is that they force a potential assignment to have at least two variables set to $0$. Combined with the requirements of the clauses $d_{(i,j,k)}$, the all-zero assignment is the only satisfying assignment.

Suppose one uses a local search algorithm on that formula. In that case, the clauses $d_{(i,j,k)}$ ``confuse'' the local search procedure because the uniform SLS procedure tends to change a variable more often from a $0$ to a $1$ than the other way around \cite{schöning2013satisfiability}. Ref.~\cite{schöning2013satisfiability} argues that the uniform version of Algorithm~\ref{alg:oracle_based_walksat} indeed achieves its worst-case runtime on 3-SAT of $O^*\left( \left( \frac{4}{3} \right) ^n\right)$.\footnote{By $O^*(a^n) = f(n)$ we mean there exists a polynomial $p(n)$ with $f(n) \leq p(n) \cdot a^n$ for $a>1$. 
Note that the algorithm discussed here differs slightly from the one we are considering, as it includes a restart after $n$ steps if no solution is found.} This runtime is indeed realized because the worst-case transition probabilities assumed in the analysis of the algorithm via a Markov chain are met by this example. Moreover, one can show that this formula is not only hard for the uniform version of Algorithm~\ref{alg:oracle_based_walksat} but also for more advanced SLS solvers that use a make- and break-count-heuristic \cite{schöning2013satisfiability}.\footnote{For the details and proofs of the above-presented statements, we refer to Chapters~5.4 and 5.7 of Ref.~\cite{schöning2013satisfiability}.}

It is interesting to ask how ``good'' an oracle must be to obtain a runtime polynomial in $n$. Let us consider oracles $O_{\{q_1, \dots ,  q_n\}}$ that output a $0$ on variable $x_i$ with probability $q_i$ for $i \in [n]$. Then, the following theorem holds about the runtime:
\begin{theo}
\label{theo:sufficiently_good_general}
    When using Algorithm~\ref{alg:oracle_based_walksat} with the oracle $O_{\{q_1, \dots ,q_n\}}$ with $\min_{i \in [n]}\{ q_i\} \geq \frac{2}{3}$ on the instance $\phi_{\text{hard}}(n)$, the expected number of steps that the algorithm takes is polynomial in $n$. To be precise, it is upper bounded by
    \begin{align}
    \mathbb{E}[\# \text{steps when using } O_{\{q_1, \dots ,  q_n\}} \mid \phi_{\text{hard}}(n)] \leq 
    \begin{cases}
         O(n^2) , & \text{if } \min_{i \in [n]}\{ q_i\} = \frac{2}{3},\\
         O(n) , & \text{if } \min_{i \in [n]}\{ q_i\} > \frac{2}{3}.
    \end{cases}
    \end{align}
\end{theo}
The intuition behind the proof, presented in Appendix~\ref{appendix:motivating_example}, is that we consider the algorithm as a random walk in one dimension. We then bound the algorithm's runtime by the expected number of steps taken in the random walk until it reaches an absorbing state, corresponding to finding a solution. The idea to study the process in this way is inspired by Refs.~\cite{Papadimitrou, schöning2013satisfiability}. For the expected number of steps in the random walk, we rely on a result from Ref.~\cite{El-Shehawey_2000}. 
Moreover, we show in Appendix~\ref{appendix:motivating_example_initialization} that addressing only the initialization step (while keeping uniform updating) in Algorithm~\ref{alg:oracle_based_walksat} offers no substantial or scalable benefit. This demonstrates that continuous access to the oracle is superior to initialization-only approaches, as the oracle would need to be inverse-polynomially close to the correct assignment on every variable to achieve a polynomial runtime.
This example explicitly constructs conditions for a ``sufficiently good'' oracle for WalkSAT on a family of provably hard instances. In Appendix~\ref{appendix:appendix:motivating_example_generalization}, we extend this idea and develop general conditions for oracle-based SLS solvers to ensure polynomial-time solutions that apply to various instances, SLS algorithms, and oracles.

\subsection{Learning oracle factories with the LLL loss}
\label{sec:learning_oracles}
So far, we have been concerned with single instances and oracles for them. Of course, in practice, we are often confronted instead with a whole class of instances that is distributed with respect to some measure $\mathcal{P}$ over the space of possible instances $\Phi$. We are interested in a solver that performs well when fed with samples from $\mathcal{P}$. As such, when using oracle-based SLS solvers, we are not primarily interested in single oracles but instead \emph{oracle factories}, which we define as functions $F$ from $\Phi$ to the space of random variables, such that, for any $\phi \in \Phi$, $F(\phi)$ is an oracle for $\phi$. 

Given a class of SAT instances, here represented abstractly by $\mathcal{P}$, and an SLS solver $S$, we can then formulate our main goal as finding 
\begin{align}
\label{eq:prob_across_instances}
    \arg \max_F \mathrm{Pr}_{\phi \sim \mathcal{P}, x \sim S(\phi, F(\phi))}(x \in \Pi(\phi)),
\end{align}
which is that specific oracle factory that maximizes the probability that $S$ returns a solution. 

We take a variational approach to this task and optimize a parameterized oracle factory $F_\theta$, where $\theta$ are some set of parameters in a parameter space $\Theta$. The total loss is a linear combination of two loss functions, described in detail below,
\begin{align}
\label{eq:total_loss}
    L(\theta, \phi) = \gamma_1 L_{\text{Gibbs}}(\theta, \phi) + \gamma_2 L_{LLL, z}(\theta, \phi),
\end{align}
   where $\gamma_1, \gamma_2 \geq 0$ are hyperparameters and the effective goal becomes to find 
\begin{align}
\label{eq:minimise_total_loss}
    \arg \min_{\theta \in \Theta} \mathbb{E}_{\phi \sim \mathcal{P}}[L(\theta, \phi)].
\end{align}
In particular, we introduce a novel loss function that is inspired by the propositions above: For a given value of $\theta$, introduce an additional parameterized functional $\mu_\theta[\phi]$ that maps an instance $\phi$ to a function from $[m]$ to the reals. We then define the \emph{Lov\'{a}sz Local Lemma} (LLL) loss as 
\begin{align}
\label{eq:LLL_loss}
    L_{LLL, z}(\theta, \phi) = \|(\max(0, \epsilon_{F_\theta(\phi), \mu_\theta[\phi]}(j)))_{j \in [m]}\|_z,
\end{align}
where $\| \cdot \|_z$ indicates the usual $p-$norm. This loss is motivated by the propositions above, which imply that a smaller loss leads to better performance of (variants of) the MT algorithm.\footnote{Indeed, a simple calculation based on the union bound lets us derive a lower bound of $1 - \mathbb{E}_{\phi \sim \mathcal{P}}[L_{LLL, 1}(\theta, \phi)]$ to~(\ref{eq:prob_across_instances}), however, since bounds based on the union bound are rarely of practical interest we refrain from stating this bound formally, since we believe the LLL loss is already sufficiently motivated.}

The second loss term\footnote{When considering the 3-SAT formula in Eq.~(\ref{eq:hard_formula}), this term (with a sufficiently large $\beta$) is indeed necessary to avoid ``bad local minima'' in the training landscape.} that we use here is the cross entropy between the thermal distribution
\begin{align}
    \gamma_{\beta, \phi}(x) = \frac{e^{-\beta \frac{ \phi(x)}{|\phi|}}}{Z_\beta(\phi)}, \quad Z_\beta(\phi) = \sum_{x \in X_n} e^{-\beta \frac{\phi(x)}{|\phi|}}
\end{align}
with respect to some inverse temperature $\beta > 0$ and the output distribution, i.e.,
\begin{align}
    L_{\text{Gibbs}}(\theta, \phi) = - \mathbb{E}_{\gamma_{\beta, \phi}} [\log(P_{F_\theta(\phi)})].
\end{align}
In Appendix~\ref{app:evaluate_loss_in_practice}, we provide details on how we evaluate this loss in practice.

\subsection{Graph neural networks as oracle factories}

In principle, many different ML models qualify as parameterized oracle factories. All that is required to train an ML model with our loss is for it to return both an oracle factory $F_\theta$ and the parameterized functional $\mu_\theta$ to evaluate the LLL loss. In our approach, we 
utilize a GNN as a deep learning model, which is a common strategy. Specifically, we use an \emph{interaction network} for this purpose~\cite{battaglia2016interaction, battaglia2018graphnetwork}.
For technical details of the GNN model and the used architecture, see Appendix \ref{app:GNN}. In particular, the output oracle is a product of independent Bernoulli distributions, one per variable. While this factorized form cannot capture correlations between variables, it enables efficient offline sampling -- a key advantage of our approach. Richer oracle families, such as autoregressive models, that capture variable correlations constitute an interesting direction for future work. 

In Appendix~\ref{app:GNN}, we also explain how SAT instances can be mapped to graphs via the common LCG representation and how the GNN's output is defined and used. At inference time, the GNN is queried once on a given instance to produce the per-variable Bernoulli parameters, which are then passed to the SLS solver and used as specified in Algorithms~\ref{alg:oracle_based_moser} and~\ref{alg:oracle_based_walksat} throughout the search without further queries to the model.

\section{Experiments}
\label{sec:experiments}
\begin{figure}
  \centering  
  \includegraphics[width=0.60\textwidth]{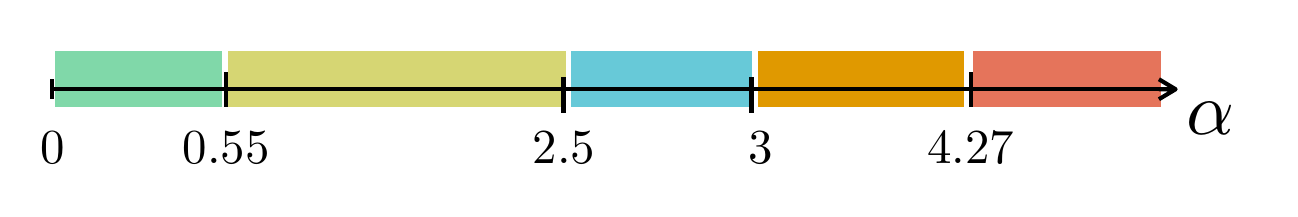}
  \caption{Simplified ``hardness regimes'' for $3$-SAT: For $\alpha < 0.55$, the uniform MT algorithm is guaranteed to solve instances efficiently (green). In practice, it efficiently solves instances up to around $2.5$ to $2.7$ (yellow)~\cite{Moser_Limit_paper}. Our experiments show that the oracle-boosted MT algorithm solves instances up to about $3.0$ to $3.2$ efficiently (blue). It remains an open question whether this cutoff can be extended to $4.27$ using better oracle factories (orange), where a phase transition occurs and instances become typically unsatisfiable (red)~\cite{phase_transition_kSAT}.}
  \label{fig:alpha_regimes}
\end{figure}
On the one hand, we train and test the above model on sets of random $3$-SAT instances of varying difficulty. Here, our measure of difficulty is the ratio $\alpha = m/n$. Random instances with lower $\alpha$ are more likely to be satisfiable and generally easier to decide. 
It is well known that random $k$-SAT undergoes a sharp phase transition from satisfiable to unsatisfiable as the clause-to-variable ratio $\alpha$ increases. Arguments from statistical physics, notably using finite-size scaling methods from spin glass theory, provide strong evidence for a $3$-SAT threshold at 
$\alpha_c(3) \approx 4.267$~\cite{phase_transition_kSAT}. Rigorous bounds place the threshold in the range 
$3.52 \leq \alpha_c(3) \leq 4.490$~\cite{kaporis_bounds3SAT, 
hajiaghayi2003_bounds3SAT, diaz2008_bounds3SAT}, and it was proven for sufficiently large $k$ that a sharp threshold 
exists~\cite{Ding2022}. We focus on the regime 
below this threshold where instances are satisfiable but become 
computationally harder as $\alpha$ increases.

Our experiments are motivated by numerical results indicating an algorithmic barrier for the original MT algorithm and WalkSAT at around $\alpha = 2.45$~\cite{Moser_Limit_paper} and $\alpha = 2.6$~\cite{Alekhnovich_2007}, respectively. 
For lower values of $\alpha$, it tends to find solutions quickly while taking exponential time for higher values of $\alpha$; see Fig.~\ref{fig:alpha_regimes} for an illustration of the regimes for the MT algorithm. We were interested in the ability of the GNN-boosted variants of these algorithms to solve instances in the ``hard'' regime. Our random 3-SAT benchmark consists of instances with $100$ to $300$ variables and a clause-to-variable ratio $1\leq \alpha \leq 4.82$. For further details about the dataset, see Appendix~\ref{app:random_3SAT_details}.

On the other hand, we train and evaluate our model on two pseudo-industrial datasets, with three difficulty levels each, made available as part of the G4SATBenchmark in Ref.~\cite{li2023g4satbench}. This benchmark includes the \emph{community attachment} (CA) model~\cite{cru2015} and the \emph{popularity-similarity} (PS) model~\cite{cru2017}. These models generate synthetic SAT instances that show statistical structures, e.g., community and locality, similar to those observed in SAT instances originating from industrial applications. The instances span three difficulty levels, containing $10$ to $400$ variables for the CA model and $10$ to $300$ variables for the PS model. Further details about the dataset are provided in Appendix~\ref{app:pseudo_industrial_details}.

Following Refs.~\cite{yolcu_2019,satenstein}, we evaluated each algorithm using three metrics: i) the average number of steps $\overline{\#}$ before finding a solution across all instances and runs, ii) the outer median (across instances) over the inner median (across runs) of steps before finding a solution, $M(\#)$, iii) the total fraction of instances solved within the cutoff time, denoted as $\%$. To accommodate multiple runs, we introduce three variants. Firstly, $\%_{\mathrm{M}}$ represents the percentage of instances solved using the median of all runs within the cutoff time. Secondly, $\%_{\mathrm{best}}$ signifies the percentage of instances solved using the best-performing run within the cutoff time. Lastly, $\%_{\mathrm{worst}}$ indicates the percentage of instances solved using the worst-performing run within the cutoff time. These variants offer distinct perspectives on the performance of the solver.

We have implemented our experiments in Python, using the JAX-framework ~\cite{jax2018github} and its graph extension JRAPH ~\cite{jraph2020github} for the GNN and optimization. The oracle-based SLS algorithms were implemented in Rust. All code and datasets used are made available as part of this publication in a GitHub repository.\footnote{\href{https://github.com/porscheofficial/sls_sat_solving_with_deep_learning.git}{https://github.com/porscheofficial/sls\_sat\_solving\_with\_deep\_learning.git}} To ensure full reproducibility, the repository includes a \texttt{readme.md} file with detailed setup instructions and a \texttt{requirements.txt} file specifying the software framework versions. 
Additionally, all model hyperparameters are documented in Appendix~\ref{app:experimental_details} (see Table~\ref{tab:hyperparams} for a consolidated overview) and provided in the repository's config files, which are thoroughly explained in the \texttt{readme.md}.

\section{Main results}
\subsection{Random 3-SAT}
We compare three different variants of both the MT algorithm and WalkSAT, namely 
\begin{itemize}
    \item the original version using the uniform oracle,
    \item a ``hybrid'' version using the trained oracle just for initialization but keeping the uniform updating, and
    \item the full-oracle algorithm using the oracle for both initialization and updating. 
\end{itemize}

In Table~\ref{tab:results}, we summarize the results when running each algorithm for up to $10^6$ steps and $1000$ runs per test instance. We find that for both solvers, the full-oracle algorithm outperforms both others across the board. Moreover, the WalkSAT algorithm dominates the MT algorithm on each variant. A particularly dramatic improvement is seen in terms of the median steps required, where the boosted MT algorithm is faster by a factor of around $8.8$, while the boosted WalkSAT provides a speed-up by a factor of around $4.6$. A less dramatic but still significant improvement is seen in the mean number of steps, with an average decrease of $33\%$ across solvers.

\begin{table}[t]
        \caption{Results from our experiments using the measures defined in Sec.~\ref{sec:experiments}. The best-performing algorithm variant is highlighted in bold.}
        \begin{tabular}[t]{lllll}
            \toprule
            Variant          & $\overline{\#}$ & $M(\#)$ & $\%_{\mathrm{M}}$ & $\alpha^*$\\
            \midrule
            uniform MT       & 3.61e5            & 1220           & 65.3 & 2.6$\pm$0.1\\ 
            ``hybrid'' MT        & 3.61e5            & 833            & 65.3 & 2.6$\pm$0.1\\ 
            full-oracle MT       & 2.25e5            & 139           & 78.6 & 3.1$\pm$0.1\\ 
            uniform WalkSAT  &  2.28e5           & 450           & 78.8 & 2.9$\pm$0.1\\ 
            ``hybrid'' WalkSAT   & 2.28e5            & 150            & 78.8 & 2.9$\pm$0.1\\ 
            full-oracle WalkSAT  & \textbf{1.62e5}   & \textbf{97}   & \textbf{84.9} & \textbf{3.2$\pm$0.1}\\ 
            \bottomrule
        \end{tabular}
        \label{tab:results}
\end{table}

\begin{figure}[b]
        \centering
        \includegraphics[width=0.65\columnwidth]{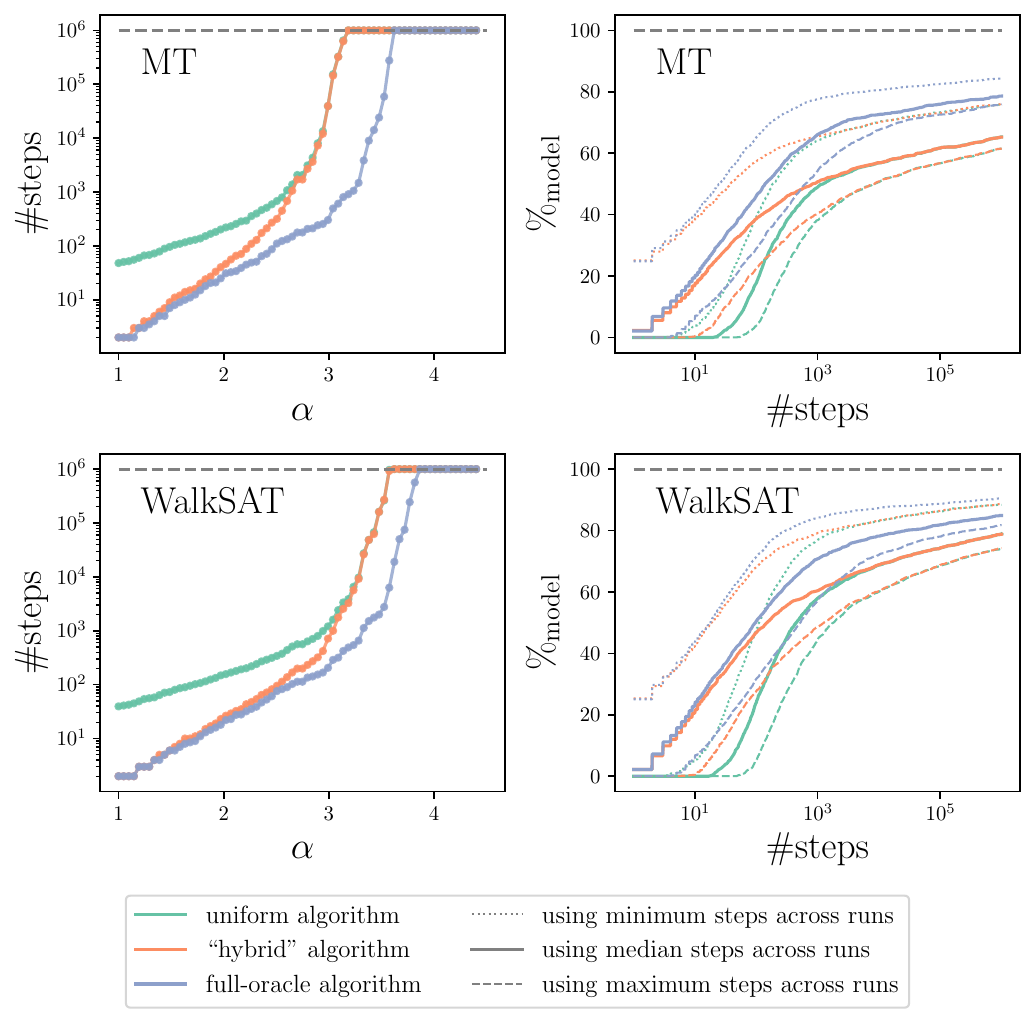}
        \captionof{figure}{Benchmark of the three described variants of the oracle-based MT algorithm (top) and oracle-based WalkSAT (bottom), respectively. Left: $M(\#)$ for respective $\alpha$. Right: $\%$ as a function of the taken steps.}
        \label{fig:initialization_vs_resampling}
\end{figure}

In Fig.~\ref{fig:initialization_vs_resampling}, we display detailed information from the experiments. Regarding $\alpha$, access to the GNN-based oracle increases the algorithmic barrier $\alpha^*$ up to $\alpha$ around $3.0$ to $3.2$ for the boosted MT algorithm. A similar behavior is found for the boosted WalkSAT algorithm. When averaging across solvers, the percentage of instances solved in the median run witnessed a $14\%$ increase for the full-oracle version compared to its uniform version. We note that the right panels of Fig.~\ref{fig:initialization_vs_resampling} display the full cumulative distribution of solved instances as a function of steps, allowing performance to be assessed at any step budget. The oracle-based solver's advantage is visible across the entire range, not only at the $10^6$ cutoff. 

Interestingly, the plots show that for both solvers, the ``hybrid'' variant that switches to uniform updating quickly loses its initial advantage over the purely uniform variant. While this variant does not tackle additional instances beyond those solved by the uniform version, it solves them in fewer steps. This showcases the advantage of giving an SLS solver continuous access to an oracle. Further results and experiments can be found in Appendix~\ref{app:random3SAT_results}. Notably, we observe that the LLL loss plays a much greater role in improving solver performance compared to the Gibbs loss. This aligns with our theoretical justification in Propositions~\ref{prop:moser} and \ref{prop:harris}, which explain why optimizing with the LLL loss leads to stronger performance. Empirical evidence supporting this claim is provided in Appendix~\ref{app:random3SAT_results}, particularly in Fig.~\ref{fig:comparison_loss_terms}.

\subsection{Pseudo-industrial dataset}
On the pseudo-industrial datasets, we train our model using only the Gibbs loss as it yields the best solvers. This also makes a lot of sense as the LLL loss is not motivated for such structures and is usually only studied in the context of random SAT. Moreover, we only evaluate our models on the oracle-version of WalkSAT as this solver is numerically stronger than the MT algorithm. 

In Table~\ref{tab:data_summary_industry_short}, we summarize the results of our experiments. Here, we benchmark the full-oracle WalkSAT algorithm against its uniform version. For each test dataset, we use that oracle for the algorithm that was trained on the corresponding training dataset. We ran each algorithm for up to $10^7$ steps and $100$ runs per test instance. As we can see, the learned oracle helps in many cases to obtain algorithms with better performance across several measures. Only for the easy datasets, there is no improvement or even a deterioration on some measures as the instances are already very simple for the uniform solver. It is striking that the relative improvement of the full-oracle WalkSAT seems to increase with the increasing difficulty of the problem, in particular when considering the instances solved. However, it also becomes clear that due to their simplicity the WalkSAT algorithm and its oracle versions are not solvers that can cope with state-of-the-art SAT solvers. Further results and a cross-benchmark can be found in Appendix~\ref{app:industry_results}. Here, we observed a notable trend: models trained on easier or medium instances often generalize well to harder instances. Surprisingly, in some cases, models trained on easier instances even outperform those trained specifically on harder ones.
\begin{table}[h]
  \centering
  \caption{
  Results from our experiments for the pseudo-industrial datasets using the measures defined in Section~\ref{sec:experiments}. The relative change $\Delta_{\mathrm{rel}}[.]$ is the difference between the metric for the uniform solver and the ML-boosted solver divided by the metric for the uniform solver. The best-performing algorithm variant is highlighted.
  }
  \begin{tabular}{lccccccc}
    \toprule
     & & $\textrm{CA}_{\textrm{easy}}$ & $\textrm{CA}_{\textrm{medium}}$ & $\textrm{CA}_{\textrm{hard}}$ & $\textrm{PS}_{\textrm{easy}}$ & $\textrm{PS}_{\textrm{medium}}$& $\textrm{PS}_{\textrm{hard}}$ \\
    \midrule
    \multirow{5}{*}{\rotatebox{90}{ML}} 
    & $\overline{\#}$ & \textbf{1.83e2} & \textbf{1.48e6} & \textbf{8.85e6} & 1.93e4 & \textbf{3.09e6} & \textbf{8.20e6}\\
    & $M(\#)$ & \textbf{7.88e1} & \textbf{5.94e3} & \textbf{1.00e7} & \textbf{6.18e1} & \textbf{6.08e4} & \textbf{1.00e7} \\
    & $\%_{\mathrm{M}}$ & \textbf{100} & \textbf{81} & \textbf{7} & 99 & \textbf{64} & \textbf{15} \\
    & $\%_{\mathrm{best}}$ & \textbf{100} & \textbf{94} & \textbf{22} & \textbf{100} & \textbf{82} & \textbf{30} \\
    & $\%_{\mathrm{worst}}$ & \textbf{100} & \textbf{73} & \textbf{5} & 99 & \textbf{53} & \textbf{11} \\
    \midrule
    \multirow{5}{*}{\rotatebox{90}{uniform}}
     & $\overline{\#}$ & 1.87e2 & 2.52e6 & 9.45e6 & \textbf{2.95e3} & 4.52e6 & 9.32e6 \\
    & $M(\#)$ & 1.18e2 & 3.52e4 & \textbf{1.00e7} & 1.47e2 & 1.82e6 & \textbf{1.00e7} \\
    & $\%_{\mathrm{M}}$ & \textbf{100} & 70 & 5 & \textbf{100} & 48 & 5 \\
    & $\%_{\mathrm{best}}$ & \textbf{100} & 85 & 10 & \textbf{100} & 68 & 14 \\
    & $\%_{\mathrm{worst}}$ & \textbf{100} & 59 & 4 & \textbf{100} & 39 & 2 \\
    \midrule
    \multirow{5}{*}{\rotatebox{90}{$\Delta_{\mathrm{rel}}[.]$ in \%}}
    & $\overline{\#}$ & -1.9 & -41.2 & -6.3 & 554.0 & -31.6 & -12.0\\
    & $M(\#)$ & -33.4 & -83.1 & 0.0 & -58.1 & -96.7 & 0.0 \\
    & $\%_{\mathrm{M}}$ & 0.0 & 15.7 & 40.0 & -1.0 & 33.3 & 200.0 \\
    & $\%_{\mathrm{best}}$ & 0.0 & 10.6 & 120.0 & 0.0 & 20.6 & 114.3 \\
    & $\%_{\mathrm{worst}}$ & 0.0 & 23.7 & 25.0 & -1.0 & 35.9 & 450.0 \\
    \bottomrule
  \end{tabular}
  \label{tab:data_summary_industry_short}
\end{table}
\section{Conclusion and future work}
This work presents a novel approach utilizing deep learning to construct oracle based SLS solvers for SAT. Beginning with a theoretical underpinning of our methodology and the notion of ``sufficiently good'' oracles, we proceed to empirically assess our approach across both random 3-SAT and pseudo-industrial instances. A primary contribution lies in the development of algorithms and associated loss functions engineered to ensure that the training process, as supported by theoretical frameworks, yields a strengthened solver, particularly evident in the realm of random SAT datasets. But also in other regimes, where no theoretical guarantee exists, the hope is that with this approach, we can create solvers for SAT and constraint optimization, more generally, that are optimized for specific applications. We have showcased this with promising results on the pseudo-industrial datasets.

The main advantage of our oracle formalism is the offline use of learned models while going beyond initialization. We have showcased this both theoretically and numerically, with good performance. However, 
it is important to acknowledge that our resulting algorithms are currently not competitive with state-of-the-art SAT solvers, and the studied instances are relatively small scale. 
This limitation is partly due to the use of the most basic SLS algorithms, which we have chosen here as a natural starting point because of the theoretical guarantees. We note that scalability to larger instances is a challenge shared by essentially all GNN-based approaches to combinatorial problems, as the computational cost of message passing grows with the size of the factor graph representation.
Nevertheless, we remain optimistic about the potential of our approach, given its general applicability and potential extension to more sophisticated SAT solvers. It may also serve as a starting point to devise more efficient quantum algorithms providing solvers for SAT
\cite{Zhang_2025, benjamin2017, boulebnane2022,schreiber2025} and QSAT \cite{cubitt2023,Gilyen}. Since quantum algorithms have the potential to approximate instances of combinatorial optimization problems well that cannot be efficiently well approximated \cite{OptimizationAdvantages,Szegedy,DecodedQuantumInterferometry,MindTheGaps}, this constitutes an interesting and promising avenue.
This establishes the promise of the oracle-based approach in crafting specialized algorithms tailored for specific problem domains. We particularly emphasize the importance of designing loss functions tailored to these specific domains. While certain aspects of our approach are left for future exploration, 
we believe the results presented in this work provide a solid foundation for 
further research in this direction.

\section*{Acknowledgments}
The authors thank Mahdi Manesh and 
Alexander Nietner for stimulating discussions. 
P.B.\ and M.K.\ thank Porsche Digital GmbH for the possibility of working on this research project. M.K.\ and J.E.\ acknowledge the support of the BMFTR (HYBRID++, QuSol, PraktiQOM), the BMWK (EniQmA), the Quantum Flagship (PasQuans2, Millenion), 
the DFG (SPP 2514), and the European Research Council (DebuQC).

\section*{Disclaimer}
The results, opinions and conclusions expressed in this publication are not necessarily those of Porsche Digital GmbH.

\bibliography{bib.bib}
\bibliographystyle{alpha}

\clearpage

\appendix

\clearpage
\appendix
\section{Detailed calculation for the motivating example of a ``sufficiently good'' oracle}
\label{appendix:motivating_example}
\begin{figure}[h]
\centering
\includegraphics[width=0.55\columnwidth]{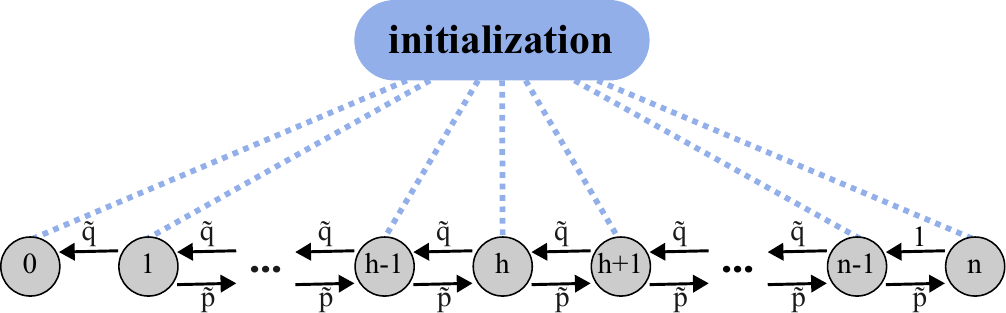}
\caption{Markov chain for a random walk in one dimension with $n+1$ states, an absorbing barrier at state $0$, and a perfectly reflecting barrier at state $n$. An initialization
procedure initializes the system in one of the states.}
\label{fig:markov_oracle}
\end{figure}

We prove Theorem~\ref{theo:sufficiently_good_general} by considering the algorithm as a random walk in one dimension (see Fig.~\ref{fig:markov_oracle}). The idea to analyze the algorithm's runtime using a Markov chain stems from Refs.~\cite{schöning2013satisfiability, Papadimitrou}, but we study it here in a broader context and generalize the setting.

We study the Markov chain depicted in Fig.~\ref{fig:markov_oracle}. In the literature, one refers to such a Markov chain as a random walk in one dimension with $n+1$ states, an absorbing barrier at state $0$, and a perfectly reflecting barrier at state $n$. Let $T_{j,k}$ be a random variable for the duration of the random walk from state $j$ to state $k$. In the end, we are interested in the expected time it takes to reach the absorbing state $0$ when starting at state $k$, which we denote as $E_k:= \mathbb{E}[T_{k,0}]$. Since the Markov chain we consider here is a well-studied object in the literature, we use the expression for $E_k$ that is derived in Ref.~\cite{El-Shehawey_2000} (see Eq.~(3.8) therein when taking the corresponding limits):
\begin{align}
\label{eq:expected_time_general}
    E_k = \begin{cases}
                \frac{1}{2\tilde{p} - 1} \left[ \left( 1 - \left(\frac{1 - \tilde{p}}{\tilde{p}} \right)^k \right)  \frac{2 (1 - \tilde{p})\tilde{p}}{2\tilde{p} - 1} \left(\frac{1- \tilde{p}}{\tilde{p}}\right)^{-n} - k \right] & \text{if} \quad \tilde{p} \neq \frac{1}{2}, \\
                2\cdot k \cdot n - k^2 & \text{if} \quad \tilde{p} = \frac{1}{2}.
            \end{cases}
\end{align}

We emphasize the existence of three different regimes. First, if $\tilde{p} < \frac{1}{2}$, the random walk is biased towards the absorbing state $0$. In this case, independent of the initialization, the expected number of steps to reach state $0$ is at most linear in $n$. Second, if $\tilde{p} = \frac{1}{2}$ the process is a symmetric random walk. In this case, independent of the initialization, the expected number of steps to reach state $0$ is at most quadratic in $n$. Third, if $\tilde{p} > \frac{1}{2}$, the random walk is biased towards the reflecting state $n$. In this case, for an arbitrary initialization, the expected number of steps is no longer polynomial in $n$.

Our final goal is to calculate how many steps the algorithm needs on expectation until it finds a solution. To illustrate why the expected algorithm's runtime can be upper bounded by the expected time it takes to reach state $0$ in the random walk in Fig.~\ref{fig:markov_oracle}, we start in a simplified setting.

Therefore, let us consider an oracle $O_q$, parameterized by $q \in [0,1]$, which independently 
outputs a $0$ with probability $q$ for each variable. It is important to note that each variable is sampled independently and identically distributed (i.i.d.). Using a simple argument, we prove the following statement, afterward generalized to Theorem~\ref{theo:sufficiently_good_general}.

\begin{theo}
\label{theo:sufficiently_good_simple}
    When using Algorithm~\ref{alg:oracle_based_walksat} with the oracle $O_q$ with $q\geq \frac{2}{3}$ on the instance $\phi_{\text{hard}}(n)$, the expected number of steps that the algorithm takes is polynomial. To be precise, it is upper bounded by:
    \begin{align}
    \mathbb{E}[\# \text{steps when using } O_q \mid \phi_{\text{hard}}(n)] \leq 
    \begin{cases}
         \frac{1}{9} n (5n -2) , & \text{if } q = \frac{2}{3} ,\\
         \frac{(2-q)\cdot (1-q)}{3q -2} \cdot n , & \text{if } q > \frac{2}{3}.
    \end{cases}
    \end{align}
\end{theo}
\begin{proof}
Let us consider Algorithm~\ref{alg:oracle_based_walksat} applied to the 3-SAT formula in Eq.~(\ref{eq:hard_formula}) with an oracle $O_q$. We are interested in the Hamming distance to the solution and keep track of this measure during our analysis. As the only satisfying assignment for this formula is the all-zero assignment $\alpha_0$, the Hamming distance of the current assignment $\alpha$ to $\alpha_0$ is simply the number of $1$'s in $\alpha$. When addressing the different clauses, the current Hamming distance $h$ changes in the following way:
\begin{itemize}[noitemsep]
    \item For clauses $d_{(i,j,k)}$:
    \begin{itemize}[noitemsep]
        \item $h \rightarrow h+1$ with probability $\tilde{p}= \frac{2(1-q)}{q + 2(1-q)}=\frac{2(1-q)}{2 - q}$,
        \item $h \rightarrow h-1$ with probability $\tilde{q}= \frac{q}{q + 2(1-q)}=\frac{q}{2-q}$, 
    \end{itemize}
    \item for clauses $c_1$ and $c_2$:
    \begin{itemize}[noitemsep]
        \item $h \rightarrow h-1$ with probability $1$.
    \end{itemize}
\end{itemize}
In the large $n$ limit, addressing clauses of the type $d_{(i,j,k)}$ dominates the process. We aim to calculate how many steps the algorithm needs on expectation until it finds a solution. For calculating this, we use the Markov chain in Fig.~\ref{fig:markov_oracle}, which upper bounds the actual process, i.e., the true Hamming distance is smaller than or equal to the state number in the Markov chain. Therefore, having reached state $0$ corresponds to having found the solution. Due to Eq.~(\ref{eq:expected_time_general}), an expression is known for the expected number of steps to reach state $0$. However, this does not yet include the initialization step. We know that the probability to initialize in state $k$ is given by
\begin{align}
\label{eq:initialization_probability}
  \operatorname{Pr}(\text{initialize in state $k$}) =\binom{n}{k} \cdot q^{n-k} \cdot (1-q)^{k}.
\end{align}
Note that we use the probability $q$ here (and not $\tilde{q}$ or $\tilde{p}$) as the initialization is done directly according to the oracle $O_q$ in contrast to the updating procedure we considered above. Therefore, the expected number of steps for the whole process (i.e., initialization and the corresponding steps until the random walk reaches state $0$) is given by
\begin{align}
    \mathbb{E}[\# \text{steps until process reaches state 0}] &= \sum_{k = 0}^n \operatorname{Pr}(\text{initialize in state $k$}) \cdot E_k.
\end{align}
Since we have
\begin{align}
    \mathbb{E}[\# \text{steps when using } O_q \mid \phi_{\text{hard}}(n)] \leq \mathbb{E}[\# \text{steps until process reaches state $0$}],
\end{align}
we can use the values calculated below as an upper bound for the runtime of our algorithm to prove the statements in Theorem~\ref{theo:sufficiently_good_simple}.

We note that at exactly $q=\frac{2}{3}$, we have $\tilde{p}=\frac{1}{2}$. Following the above elaboration about the different regimes for the random walk, we will address the cases (i) $q > \frac{2}{3}$ and (ii) $q = \frac{2}{3}$ independently when including the initialization step. In the first case, we consider a random walk biased towards the absorbing state $0$ (as $q>\frac{2}{3}$ implies $\tilde{p}<\frac{1}{2}$ in the random walk) and in the latter case a symmetric random walk (as $q=\frac{2}{3}$ implies $\tilde{p}=\frac{1}{2}$ in the random walk). If $q < \frac{2}{3}$, the random walk is biased towards the reflecting state $n$. Therefore, $q=\frac{2}{3}$ constitutes a boundary for the WalkSAT algorithm on this $3$-SAT instance.

\subsection{Case (i) where $q > 2/3$}
For the first case, where $q > \frac{2}{3}$ and thus $\tilde{p} < \frac{1}{2}$, we have after rewriting $E_k$ such that all the terms are positive for $\tilde{p}<\frac{1}{2}$
\begin{align}
\begin{split}
    & \mathbb{E}[\#\text{steps until process reaches state $0$}] \\
    &= \sum_{k = 0}^n \operatorname{Pr}(\text{initialize in state $k$}) \cdot E_k\\
    & = \sum_{k = 0}^n \left[\binom{n}{k} q^{n-k} (1-q)^{k}\right] \cdot 
    \left[\frac{k}{1 - 2\tilde{p}} - \frac{2(1- \tilde{p})\tilde{p}}{(1 - 2\tilde{p})^2} \left(\frac{\tilde{p}}{1- \tilde{p}}\right)^{n} \left(\left(\frac{1 - \tilde{p}}{\tilde{p}} \right)^k - 1\right) \right] \\
    & = \frac{1}{1 - 2\tilde{p}} \sum_{k = 0}^n \left[\binom{n}{k} q^{n-k} (1-q)^{k} \cdot k\right]\\
    & \quad - \frac{2(1- \tilde{p})\tilde{p}}{(1 - 2\tilde{p})^2} \left(\frac{\tilde{p}}{1- \tilde{p}}\right)^{n} \sum_{k = 0}^n \left[\binom{n}{k} q^{n-k} (1-q)^{k} \cdot  \left(\left(\frac{1 - \tilde{p}}{\tilde{p}} \right)^k - 1\right) \right] \\  
    & = \frac{1}{1 - 2\tilde{p}} \left[n(1-q)\right] - \frac{2(1- \tilde{p})\tilde{p}}{(1 - 2\tilde{p})^2} \left(\frac{\tilde{p}}{1- \tilde{p}}\right)^{n} \left[\frac{(2 q \tilde{p} - q - \tilde{p} + 1)^n}{\tilde{p}^n} - 1\right] \\  
    & = \frac{(2-q)(1-q)}{3q - 2} n - \frac{2 (1-q) q}{(3q - 2)^2} \left(\frac{2(1-q)}{q} \right)^n \left[ \left(\frac{3q}{2}\right)^n - 1\right].
\end{split}
\end{align}
Here, we have inserted the definition of $\tilde{p}$ in the second-last step. Since the second term in the last line is positive for $q>\frac{2}{3}$, 
we have, in particular,
\begin{align}
\label{eq:ineq_processes}
    \mathbb{E}[\#\text{steps until process reaches state $0$}] \leq \frac{(2-q)(1-q)}{3q - 2} n,
\end{align}

which is one of the results stated in Theorem~\ref{theo:sufficiently_good_simple}.

\subsection{Case (ii) where $q = 2/3$}
For the second case, where $q = \frac{2}{3}$, we proceed in a similar way and find
\begin{align}
\begin{split}
    &\mathbb{E}[\# \text{steps until process reaches state $0$}]\\
    &= \sum_{k = 0}^n \operatorname{Pr}(\text{initialize in state $k$}) \cdot E_k\\
    & = \sum_{k = 0}^n \left[\binom{n}{k} q^{n-k} (1-q)^{k}\right] \cdot 
    \left[2\cdot k \cdot n - k^2 \right] \\
    & = 2\cdot n \cdot \left[ \sum_{k = 0}^n \binom{n}{k} q^{n-k} (1-q)^{k} \cdot k\right] -  \left[\sum_{k = 0}^n \binom{n}{k} q^{n-k} (1-q)^{k} \cdot k^2 \right] \\
    & = 2\cdot n \cdot \left[n(1-q) \right] -  \left[n \left(n \left(1 - q\right) - q\right) \left(1 - q\right) \right]\\
    & = \frac{1}{9} n (5 n - 2) \\
\end{split}
\end{align}
where we have inserted $q = \frac{2}{3}$ in the last step. As this is the other result stated in Theorem~\ref{theo:sufficiently_good_simple}, this concludes the proof.
\end{proof}

\subsection{Generalization to Theorem~\ref{theo:sufficiently_good_general}}
Below, we show that Theorem~\ref{theo:sufficiently_good_simple} can be generalized to Theorem~\ref{theo:sufficiently_good_general}.
\begin{proof}
    It should be clear that whenever we use oracles of the type $O_{\{q_1, \dots ,  q_n\}}$ (i.e., oracles that output a $0$ on variable $x_i$ with $q_i$), if $\min_{i\in[n]}\{ q_i\} \geq \frac{2}{3}$ the true process is still upper bounded by the one considered above when replacing $q$ by $\min_{i\in[n]} q_i$. Therefore, we can use the abovementioned techniques and bounds to prove the generalized statement in Theorem~\ref{theo:sufficiently_good_general}.
\end{proof}

\subsection{Initialization-only approach}
\label{appendix:motivating_example_initialization}
In this paragraph, we would like to address what happens if the oracle is only used for the initialization, but the uniform updating is kept. We argue here that, asymptotically for large $n$, $\mathbb{E}[\# \text{steps until process reaches state $0$}]$ and $\mathbb{E}[\# \text{steps when using } O_q \mid \phi_{\text{hard}}(n)]$ show the same asymptotic scaling as there are only the two clauses $c_1$ and $c_2$ that lead to a different random walk but $3 \cdot \binom{n}{3}$ clauses of type $d_{(i,j,k)}$ which lead to the described random walk. Therefore, for a large enough $n$, the difference between the true and studied process is negligible, and we can use the results from above. Indeed, one can show that the true transition probability to the right $\tilde{p}_{\text{true}}$ is given by
\begin{align}
    \tilde{p}_{\text{true}} = \frac{3 \binom{n}{3} \cdot \tilde{p} + 2 \cdot 0}{3 \binom{n}{3} + 2},
\end{align}
and thus 
\begin{align}
\tilde{p}_{\text{true}} = \tilde{p} - \Theta(n^{-3}).
\end{align}
However, we stick to $\tilde{p}$ here because this is only a heuristic argument. If stick to the uniform updating when using Algorithm~\ref{alg:oracle_based_walksat}, we have $\tilde{p} = \frac{2}{3}$ in the random walk. Following Eq.~(\ref{eq:expected_time_general}), 
we find
\begin{align}
    E_k = 4\cdot 2^n \left( 1 - \left(\frac{1}{2}\right)^k \right) - 3k.
\end{align}
As we keep the oracle $O_q$ for initialization, the probability to initialize in state $k$ is still given as in Eq.~(\ref{eq:initialization_probability}). Thus, after doing the math
\begin{align}
    \mathbb{E}[\# \text{steps until process reaches state $0$}] & = \sum_{k = 0}^n \operatorname{Pr}(\text{initialize in state $k$}) \cdot E_k\\
    & = 4 \cdot 2^n \left( 1 - \left( \frac{q}{2} + \frac{1}{2} \right)^n \right) - 3 n (1-q).
\end{align}
If our oracle is exponentially close to a perfect one, i.e., $q \geq 1 - 2^{-(n-1)}$, the calculated runtime is asymptotically linear in $n$. In all other cases, the runtime remains exponential in $n$. 

A strategy that works better than using this algorithm would be to ``resample the whole assignment'' until a solution is found. If we sample according to the oracle, an all-zero assignment is found with probability $q^n$. Therefore, we expect $\frac{1}{q^n}$ tries until we sample an assignment. It should be clear that even in this case, it remains hard to come by with an oracle that is ``sufficiently good'' as this would require an oracle inverse-polynomially close to the optimal one.

\subsection{Generalization to further algorithms and instances}
\label{appendix:appendix:motivating_example_generalization}
One can generalize this approach to other solvers and also to other instances. To find oracle-based SLS algorithms that return a solution to an instance $\phi$ in a runtime that is polynomial in the number of variables, we formulate conditions on an oracle.

Consider a satisfiable SAT instance $\phi$ in CNF with $n$ variables and $m$ clauses. We fix one satisfying assignment $\alpha^* \in \Pi(\phi)$. This analysis focuses on finding exactly this $\alpha^*$. As it is a solution, there is at least one literal in every clause $c \in \phi$ that receives truth value $1$ under assignment $\alpha^*$. For any assignment $\alpha \in \{0,1\}^n$, let us denote by $h(\alpha, \alpha^*)$ the Hamming distance of the assignment $\alpha$ to the solution $\alpha^*$. We consider SLS algorithms in their simplest form as we introduced them in Section~\ref{sec:oracle_based_solvers}, where an oracle $O$ may only used in the algorithms during the \texttt{initialize} and the \texttt{update} routines and not for choosing the violated clause that is addressed next or anything else. If the update routine addresses a clause $c$, it can only change the assignment of the variables in $c$ while leaving the others unchanged. Suppose an assignment $\alpha$ violates clause $c = ( l_1 \vee l_2 \vee \dots \vee l_k )$. In that case, we can deduct the current assignment of variables on the variables in $c$, i.e., $\alpha|_{V(c)} = \{\bar{l}_1, \bar{l}_2, \dots, \bar{l}_k\}$ because a clause explicitly excludes one assignment if the formula is in CNF. Let $\alpha_{\mathrm{updated}}$ be the updated assignment when addressing clause $c$ with the update routine of the oracle-based SLS solver $S$ with oracle $O$. As the update procedure is generally probabilistic, this can be seen as a random walk with certain probabilities. Therefore, let $\tilde{q}_{\mathrm{success}}(c|S, O, \alpha^*)$ be the probability that this update step produces an assignment $\alpha_{\mathrm{updated}}$ such that:
\begin{align}
    h(\alpha_{\mathrm{updated}}, \alpha^*) \leq h(\alpha, \alpha^*) - 1,
\end{align}
e.g., the probability that by applying the update step, the Hamming distance to the solution decreases by at least $1$. 

In the following theorem, we formulate conditions on oracle-based SLS algorithms such that they return a solution in linear time when addressing an instance $\phi$. For this theorem, we restrict ourselves to update routines that change at most one variable per update step.

\begin{theo}
    Let $\phi$ be a satisfiable SAT instance in CNF with $n$ variables and $m$ clauses. We fix an oracle-based SLS solver $S$ (as defined above) that uses an oracle $O$. We require to have an update routine that changes at most one variable in one update step. If there is a fixed satisfying assignment $\alpha^* \in \Pi(\phi)$ such that
    \begin{align}
        \tilde{q}_{\mathrm{success}}(c | S, O, \alpha^*) > \frac{1}{2} \quad \forall c \in \phi,
    \end{align}
    the resulting oracle-based SLS solver $S$ with oracle $O$, applied to $\phi$ returns a solution in at most $O(n)$ steps.

    Moreover, if there is a fixed satisfying assignment $\alpha^* \in \Pi(\phi)$ such that
    \begin{align}
        \tilde{q}_{\mathrm{success}}(c | S, O, \alpha^*) 
        \geq \frac{1}{2} \quad \forall c \in \phi,
    \end{align}
    the resulting oracle-based SLS solver $S$ with oracle $O$, applied to $\phi$ returns a solution in at most $O(n^2)$ steps.
\end{theo}

\begin{proof}
    We sketch a proof here as this is a simple corollary from the formalism discussed above. As the update routine changes at most one variable per update step, the Hamming distance to the solution can change at most by $1$ in one update step. Consider the Markov chain described in Fig.~\ref{fig:markov_oracle} and the expression for the expected time to reach state $0$ in Eq.~(\ref{eq:expected_time_general}). As usual, the state number is an upper bound for the Hamming distance of the current assignment to the solution $\alpha^*$. If $\tilde{q}_{\mathrm{success}}(c | S, O, \alpha^*) > \frac{1}{2}$ for all $c \in \phi$, then essentially the overall transition probability to go from state $i$ to state $i-1$ is greater than $\frac{1}{2}$ (corresponding to $\tilde{p} = (1 - \tilde{q}) < \frac{1}{2}$ in Eq.~(\ref{eq:expected_time_general})). Therefore, on expectation, the SLS algorithm returns a solution, on expectation, in at most $O(n)$ steps.
    Therefore, even if we initialize in state $n$, the process reaches state $0$, on expectation, in at most $O(n)$ steps. 
    
     In the second case, i.e., $\tilde{q}_{\mathrm{success}}(c | S, O, \alpha^*) \geq \frac{1}{2}$, the problem reduces to the analysis of a symmetric random walk. Therefore, following the expression in Eq.~(\ref{eq:expected_time_general}), the algorithm returns a solution, on expectation, in at most $O(n^2)$ steps.
\end{proof}
When considering any $2$-SAT instance and the WalkSAT algorithm with the uniform oracle $I$, we note that the theorem above shows the quadratic runtime of the uniform WalkSAT algorithm on $2$-SAT, as $\tilde{q}_{\mathrm{success}}(c | \mathrm{WalkSAT}, I, \alpha^*) \geq \frac{1}{2}$ for any clause $c$. This fact is not surprising but more a sanity check, as our analysis essentially breaks down to the argument from Ref.~\cite{Papadimitrou}.
\section{Evaluating the loss in practice}
\label{app:evaluate_loss_in_practice}
In Sec.~\ref{sec:learning_oracles} of the main text, we motivated the goal to find an oracle factory that minimizes the total loss in Eq.~(\ref{eq:total_loss}). However, in practice, we do not have access to $\mathcal{P}$. As is common practice, we instead estimate the loss using samples from a training set $\mathcal{X} = { (X_i,\phi_i) }_{i=1}^N$ of $N$ instances that we assume to be sampled independently from $\mathcal{P}$. Here, $X_i=\{ x^{i,j} \}_{j=1}^{N_i}$ describes a set of unique $N_i$ assignments for the problem instance $\phi_i$. For each assignment $x^{i,j}$, let $e^{i,j}$ denote the number of clauses it violates. For every instance, we can calculate LLL loss by first explicitly constructing the dependency graph $\mathcal{D}_{F_\theta(\phi_i)}(\phi_i)$. This is relatively simple in our case because, due to the simple Bernoulli structure of our output oracles, we know that the neighborhood of a given clause node $j$ is simply the set of clause nodes whose corresponding variable sets intersect with that of $c_j$, i.e., $\Gamma(j):= \{ j' \in [m] | j' \neq j, V(c_{j'}) \cap V(c_j) \neq \emptyset \}$. Given a batch $B \subset [|\mathcal{X}|]$ of instances, we estimate the LLL loss by calculating the $z$-norm of the whole loss vector, creating the dependency graph of the batch as the union of the dependency graphs of batch elements. 

Regarding the Gibbs loss, we follow Ref.~\cite{Deepmind2020} and generate, for a given batch $B$, an estimator
\begin{align}
    \hat{L}_{\text{Gibbs}}(\theta) = - \sum_{i \in B} \sum_j^{N_i} \omega_{ij} \log{(P_{F_\theta(\phi_i)}(x^{i,j}))}
\end{align}
with the Gibbs weights
\begin{align}
    \omega_{ij} = \frac{\exp{\left(-\beta \frac{\phi_i(x^{i,j})}{|\phi_i|}\right)}}{\sum_{k=1}^{N_i} \exp{\left(-\beta \frac{\phi_i(x^{i,k})}{|\phi_i|}\right)}}.
\end{align}

\section{Details of the GNN architecture}
\label{app:GNN}
In the following, we consider the technicalities of the \emph{InteractionNetwork}~\cite{battaglia2016interaction, battaglia2018graphnetwork} we have used in this work. Like all other GNNs, this type of network maps graphs to graphs. Consider a graph $G=(N, E)$, with a set of nodes $N = \{\mathbf{n}_i\}_{i=1:|N|}$ and a set of edges $E= \{(\mathbf{e}_k, r_k, s_k)\}_{k=1:|E|}$, where $\mathbf{n}_i \in \mathbb{R}^{d_N}$ are feature vectors of the nodes, for some $d_N \in \mathbb{N}$, $\mathbf{e}_k \in \mathbb{R}^{d_E}$ are feature vectors of the edges, for some $d_E \in \mathbb{N}$, and $r_k, s_k \in [|N|]$ are indices of the receiver and source nodes for the $k$-th edge, respectively. A single layer of the \emph{InteractionNetwork} (IN) takes $G$ as input and updates both edge and node features based on a message-passing algorithm whose pseudo-code is shown as Algorithm~\ref{alg:gn}. This algorithm requires the specification of a node update function $\eta^N$, an edge update function $\eta^E$, and an edge aggregation function $\rho^E$. Note that, depending on the update and aggregation functions, the dimensions of the features in $G'$ might differ from those of $G$.

\begin{algorithm}[H]
\begin{algorithmic}[1]
    \FOR {$k\in \{1\ldots{}|E|\}$}
        \STATE $\mathbf{e}_k^\prime\gets \eta^E\left(\mathbf{e}_k, \mathbf{n}_{r_k}, \mathbf{n}_{s_k}\right)$
    \ENDFOR
    \FOR {$i\in \{1\ldots{}|N|\}$}
        \STATE \textbf{let} $E'_i = \left\{\left(\mathbf{e}'_k, r_k, s_k \right)\right\}_{r_k=i,\; k=1:|E|}$
        \STATE $\mathbf{\bar{e}}'_i \gets \rho^E\left(E'_i\right)$
        \STATE $\mathbf{n}'_i \gets \eta^N\left(\mathbf{\bar{e}}'_i, \mathbf{n}_i, \right)$
    \ENDFOR
    \STATE \textbf{let} $N' = \left\{\mathbf{n}'\right\}_{i=1:|N|}$
    \STATE \textbf{let} $E' = \left\{\left(\mathbf{e}'_k, r_k, s_k \right)\right\}_{k=1:|E|}$
    \RETURN $(E', N')$
\end{algorithmic}
\caption{A single IN layer}
\label{alg:gn}
\end{algorithm}

Here, we use as update functions $\eta^N$ and $\eta^V$ simple neural networks of the form 
\begin{align}
    \eta = LN \circ RL \circ FL_{d_u} \circ RL \circ FL_{d_{u - 1}} \circ \dots \circ FL_{d_{1}},
\end{align}
where $LN$ is a layer normalization layer \cite{ba2016layer}, $RL$ is a rectified linear unit layer~\cite{agarap2018relu}, and $FL_d$ is a fully connected layer from an input feature dimension to an output dimension $d$. Hence, specifying an update function in our case requires a list $(d_t)_{t=1}^u$ of layer dimensions, one for each of the $u$ layers. Here, we use a summation over the elements of $E_i'$ as an edge aggregation function. 

A full application of an IN consists of the consecutive application of several single IN layers, followed by a fully connected layer $FL_1$ to ensure that outgoing features across nodes and edges are one-dimensional. That is, for an IN consisting of $l$ layers, the output of the IN is 
\begin{align}
    IN_\theta(G) = FL_1 \circ IN_{\theta_l} \circ IN_{\theta_{l-1}} \circ \dots IN_{\theta_1}(G).
\end{align}
Here, $\theta_r$ denotes the parameter vector for the $r$-th layer, whose values specify the entries of the fully connected layers in its update functions. We have $\theta = (\theta_{FL}, \theta_l, \theta_{l-1}, \dots, \theta_1)$, where $\theta_{FL}$ are the parameters in the last layer.

\subsection{Representing SAT instances as graphs}
\begin{figure}
    \centering
    \includegraphics[width=0.45\columnwidth]{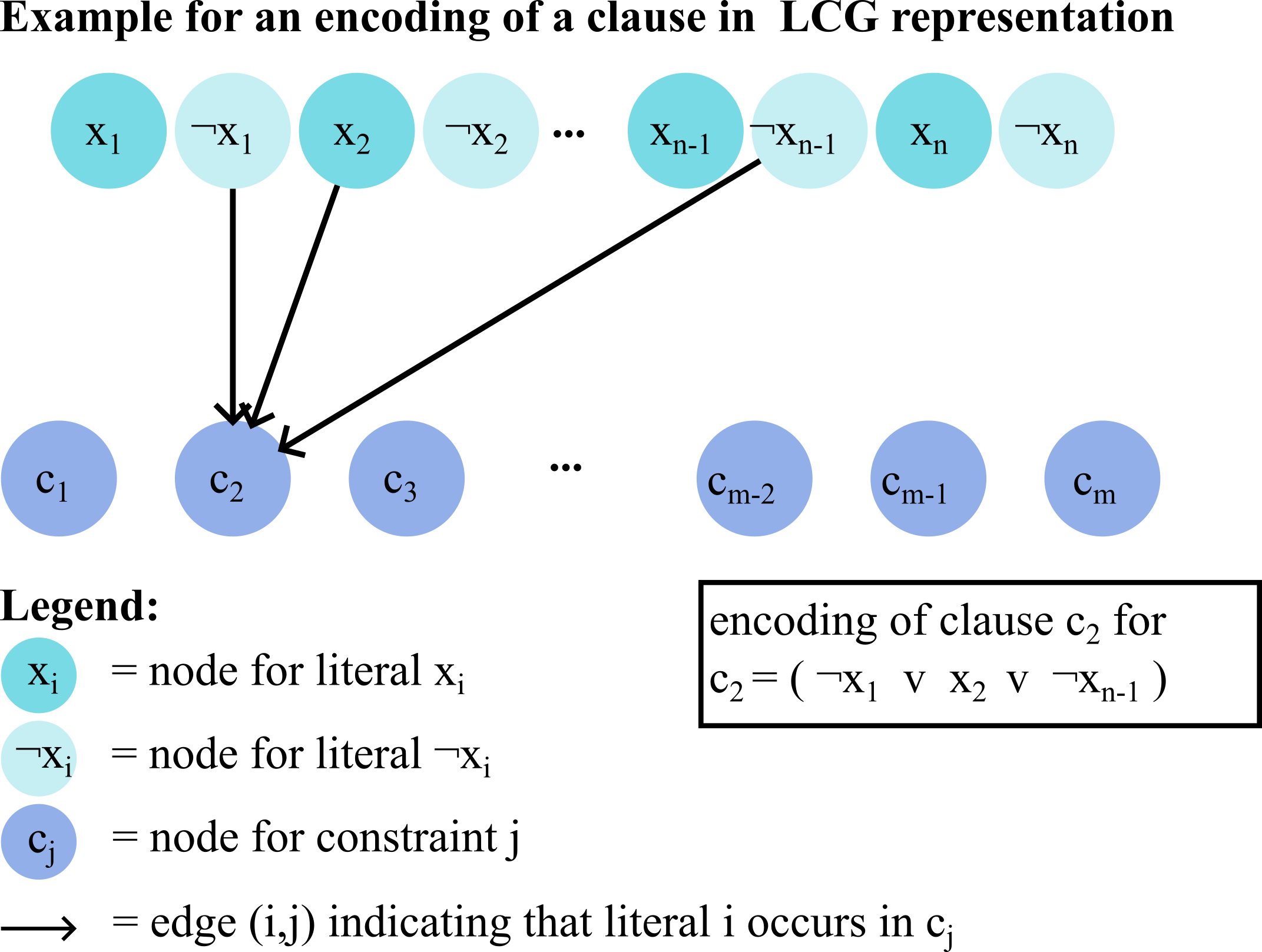}
    \caption{Example for encoding a single clause into LCG representation.}
    \label{fig:sat_encoding}
\end{figure}

Since GNNs map graphs to graphs, to feed a SAT instance into a Graph network, we need to represent it as a graph. Fortunately, there exist various simple graph representations of Boolean formulas. We use the common LCG representation:\footnote{In the code accompanying this work, we also implemented the architecture below using the alternative VCG representation. For the sake of brevity, here we only present the LCG representation, which yielded better results.} Let $\phi$ be a SAT formula in CNF form with $n$ variables $(v_i)_{i=1}^n$ and $m$ clauses $(c_j)_{j=1}^m$. We introduce a directed tripartite graph $G = (N, E)$, consisting of one set $N_C$ of $m$ nodes, one per constraint, as well as two sets $N^+_V$ and $N^-_V$ of $n$ nodes each, one per variable. The initial node embeddings $\mathbf{n}$ are one-hot encodings for which of these three sets a given node belongs. The edge set $E$ further consists of two groups. The first group connects the $i$-th node in $N^+_V$ with the $i$-th node in $N^-_V$, for $i \in [n]$. The second group connects, for each clause $c_j \in \phi$ and every variable $v_i \in V^+(c_j)$, the $j$-th node in $V_C$ with the $i$-th node in $N^+_V$, and also for every variable $v_i \in V^-(c_j)$, the $j$-th node in $V_C$ with the $i$-th node in $N^-_V$. The initial edge embeddings $\mathbf{e}$ are, again, simply one-hot encodings of which of these two groups a given edge falls into. Fig. \ref{fig:sat_encoding} contains a graphical depiction of the LCG representation. 

\subsection{Defining the GNN's output}
Let $IN_\theta(\phi)$ denote the output of the IN when feeding an instance $\phi$ into it using the above LCG representation. Recall that we require an oracle $F_\theta(\phi)$ as well as the functional $\mu_\theta(\phi)$ as output from the GNN in order to evaluate the loss~(\ref{eq:total_loss}). Due to the final layer, the output features across all nodes are one-dimensional. Let $\mathbf{n'}^C$ denote the vector of outgoing node features for the $m$ clause nodes, and similarly $\mathbf{n'}^+$ and $\mathbf{n'}^-$ for the $n$ nodes in $N^+_V$ and $N^-_V$ respectively.
We then define the oracle $F_\theta(\phi)$ the random variable over $\{0,1\}$ with measure 
\begin{align}
    P_{F_\theta(\phi)}(x) = \Pi_i \left(\tilde{w}_i^{x_i} (1 - \tilde{w}_i)^{(1- x_i)}\right),
\end{align}
where 
\begin{align}
    \tilde{w}_i = \frac{\mathbf{n'}^+_i}{\mathbf{n'}^+_i + \mathbf{n'}^-_i}.
\end{align}
In other words, the output oracle generates an assignment $x$ simply by sampling the value of the $i$-th variable as a Bernoulli trial with success probability $\tilde{w}_i$ and does so independently for each variable.
We further define the functional $\mu_\theta[\phi]$ via the mapping
\begin{align}
    \mu_\theta[\phi](j) = \frac{\hat{w}_j}{1 - \hat{w}_j},
\end{align}
where
\begin{align}
    \hat{w}_j = \mathrm{sigmoid}(\mathbf{n'}^C_j).
\end{align}
The reasoning behind this somewhat unintuitive mapping may best be understood via an alternative formulation of the LLL. Please see Ref.~\cite{MoserPhD} for details.
\section{Experimental details}
\label{app:experimental_details}

\begin{table}[h]
\centering
\caption{Summary of hyperparameters used across all experiments.}
\label{tab:hyperparams}
\begin{tabular}{
        l
        c
        c
    }
\toprule
        \textbf{Hyperparameter} 
        & \textbf{Random 3-SAT} 
        & \textbf{Pseudo-industrial} \\
        \midrule
        Message passing steps $l$        & 5         & 5 \\
        Layer dimension $d$              & 200       & 200 \\
        Number of layers per MLP         & 2         & 2 \\
        LLL norm $z$                     & 2         & 2 \\
        Inverse temperature $\beta$      & $10^{6}$  & $10^{6}$ \\
        $\gamma_1$ (Gibbs weight)        & 1         & 1 \\
        $\gamma_2$ (LLL weight)          & 1         & 0 \\
        Initial learning rate            & $10^{-1}$ & $10^{-1}$ \\
        Final learning rate              & $10^{-3}$ & $10^{-3}$ \\
        Batch size                       & 1         & 1 \\
        Training epochs                  & 200       & 1000 \\
        Training instances               & 396       & 800 (easy \& medium), 140 (hard) \\
        Runs per test instance           & 1000       & 100 \\
        Step cutoff                      &  $10^{6}$  & $10^{7}$ \\
        \bottomrule
\end{tabular}
\end{table}

\subsection{Random 3-SAT}
\label{app:random_3SAT_details}
We have trained our model on a dataset of $396$ satisfiable instances equally distributed between $n \in [100,200,300]$ and $1 \leq \alpha \leq 4.82$. We have generated these instances with the python library CNFgen (see Ref.~\cite{cnfgen_lib}) and post-selected on satisfiable instances. We have used the Glucose3 solver~\cite{glucose} implemented in the 
\emph{Pysat library} in Ref.~\cite{pysat} to generate the solutions. As it turns out, this number of training instances is already enough for our model to learn the important characteristics of these SAT instances. For the \emph{interaction network}, we have used $l=5$ message passing steps and two layers of dimension $d = 200$ for every fully connected layer in every message passing step. As hyperparameters for the loss, we used $z=2, \beta = 10^6, \gamma_1 = 1, \gamma_2 = 1$. For the optimization, we used the ADAM-optimizer with a dynamical learning rate decaying from an initial learning rate of $10^{-1}$ exponentially to a final learning rate of $10^{-3}$, a batch size of $1$, and $200$ training epochs.

To evaluate the resulting model, we ran the oracle-based WalkSAT and MT algorithms on an evaluation set of $2052$ instances, also containing an equal number of instances across $n \in [100,200,300]$ and $1 \leq \alpha \leq 4.82$. We ran each algorithm for up to $10^6$ steps and $1000$ runs per test instance.

\subsection{Pseudo-industrial datasets}
\label{app:pseudo_industrial_details}
As pseudo-industrial datasets, we have used the ones provided in the benchmark from Ref.~\cite{li2023g4satbench}. For generating these instances, the Community Attachment (CA) model from Ref.~\cite{cru2015} and the 
\emph{Popularity-Similarity} (PS) model from Ref.~\cite{cru2017} have been used. These generators aim to generate synthetic SAT instances that show similar statistical features as SAT instances from real-world applications. In general, the CA model was designed to mimic the community structures and modularity features that are observed in real-world instances \cite{cru2015, li2023g4satbench}, whereas the PS model samples instances that show, on the one hand, a power-law distribution in the number of variable occurrences in the formula (popularity) but also a good clustering between them (similarity) \cite{cru2017, li2023g4satbench}.

The benchmark contains three different difficulty levels for each model, with increasing difficulty from easy to medium to hard. For both the CA and the PS model, the easy dataset contains formulas with $10$ to $40$ variables; the medium dataset contains instances with $40$ to $200$ variables. The hard CA dataset consists of instances with $200$ to $400$ variables, and the hard PS dataset consists of instances with $200$ to $300$ variables. For the precise parameters to choose in the generators, we refer to Ref.~\cite{li2023g4satbench} as we used precisely the ones therein. 

For our dataset, we post-selected the satisfiable instances and used the Glucose3 solver~\cite{glucose} implemented in the Pysat library in Ref.~\cite{pysat} to generate the solutions. When training on the easy and medium datasets, we used $800$ instances. In the case of the hard dataset, we trained on $140$ instances. To evaluate the model, we used for all levels of difficulty $100$ satisfiable instances.

As for the 3-SAT dataset we used in the \emph{interaction network}, $l=5$ message passing steps and two layers of dimension $d = 200$ for every fully connected layer in every message passing step. As hyperparameters for the loss, we used $z=2, \beta = 10^6, \gamma_1 = 1, \gamma_2 = 0$. For the optimization, we used the ADAM-optimizer with a dynamical learning rate decaying from an initial learning rate of $10^{-1}$ exponentially to a final learning rate of $10^{-3}$, a batch size of $1$, and $1000$ training epochs. We did not use the LLL Loss for the industrial datasets as we found that it did not increase the performance of the resulting solvers. This also makes a lot of sense, as the LLL is usually studied in the context of random SAT instances.

\subsection{Experimental details for practitioners}
While experimenting with hyperparameters, we have learned that the model leads to the best results when using a batch size of $1$. We have not experimented too much with other hyperparameters. Thus, we are convinced that tuning the hyperparameters can further boost the performance of the pre-trained model.

We ran the experiments on the CPU of an HP Z8 G4 workstation equipped with 256 GB RAM and an Intel® Xeon® Gold 6130 x 64 processor, running Debian GNU/Linux 12 (bookworm). All the reported experiments can be reproduced using the configuration files in the accompanying GitHub repository. For all our datasets, we created for each instance within a dataset $500$ candidates that are used in the Gibbs loss. We created them by randomly flipping $m \sim \mathrm{Uniform}(0,0.3\cdot n)$ variable assignments in the solution string. Here, $n$ is the number of variables, as usual.

\section{Detailed numerical results for the experiments}
\subsection{Random 3-SAT}
\label{app:random3SAT_results}

\begin{figure}
    \centering
    \includegraphics[width=0.65\textwidth]{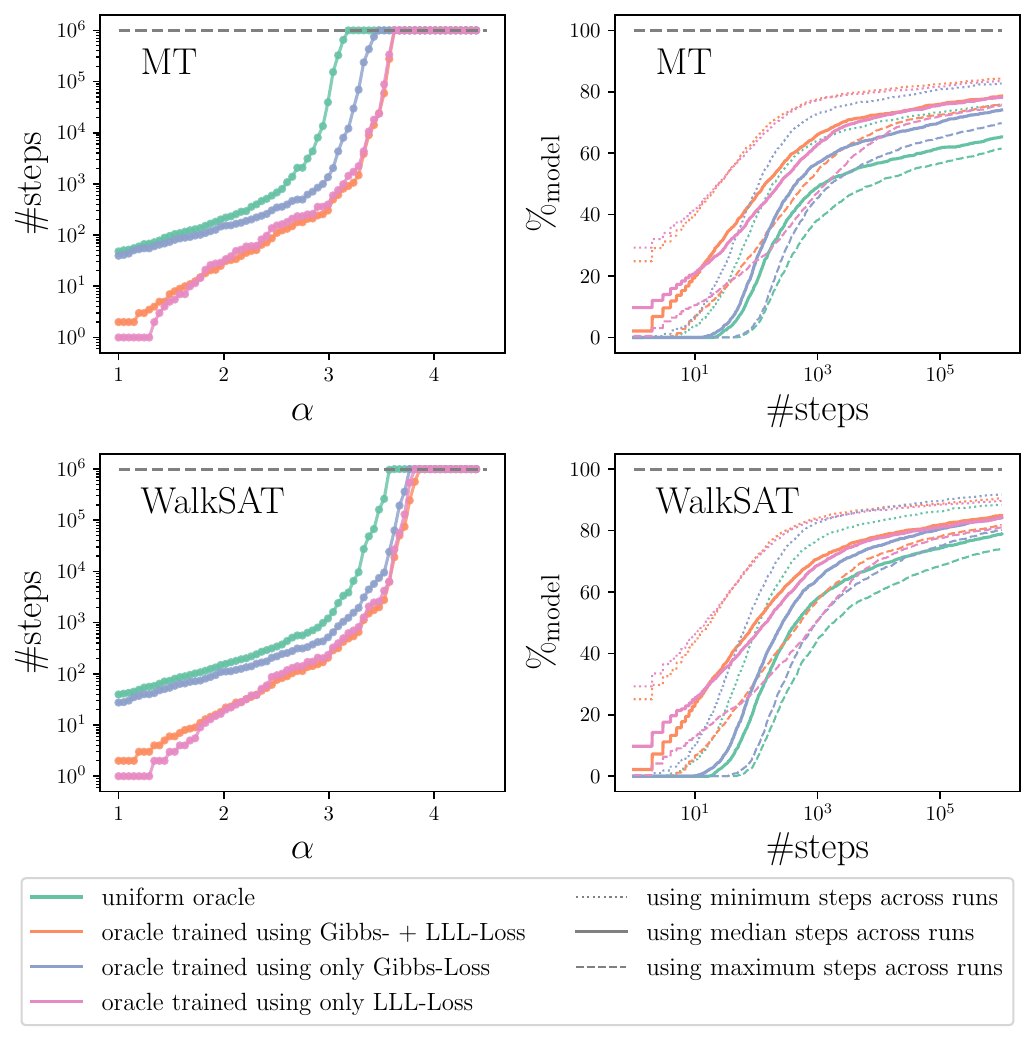}
    \caption{Comparison of the MT algorithm (top) and WalkSAT (bottom) when switching on and off the individual loss terms in the training of the underlying model. The uniform versions are plotted as a benchmark. Left: $M(\#)$ for respective $\alpha$. Right: The different measures for $\%$ as a function of the taken steps.}
    \label{fig:comparison_loss_terms}
\end{figure}

Since our loss consists of two terms (that we weigh equally in the experiments), an obvious question to ask is whether both of them contribute to the improvement in the performance of the algorithm. In Fig.~\ref{fig:comparison_loss_terms} we plot the same statistics as done in the main text above, comparing the uniform variant with the full boosted variant, with the model being trained on only one, the other, and both loss terms. Overall, we find that the combination of the two loss terms produces the best results. However, the Lov\'{a}sz Local Lemma loss contributes significantly more to the improvement than the Gibbs loss. This is especially surprising for the oracle-based WalkSAT since there is no theoretical result or argument known to the authors that would connect the WalkSAT algorithm to the Lov\'{a}sz Local loss or the conditions in Propositions \ref{prop:moser} and \ref{prop:harris}. 

\subsection{Pseudo-industrial datasets}
\label{app:industry_results}
\begin{figure}
    \centering
    \includegraphics[width = 0.45 \columnwidth]{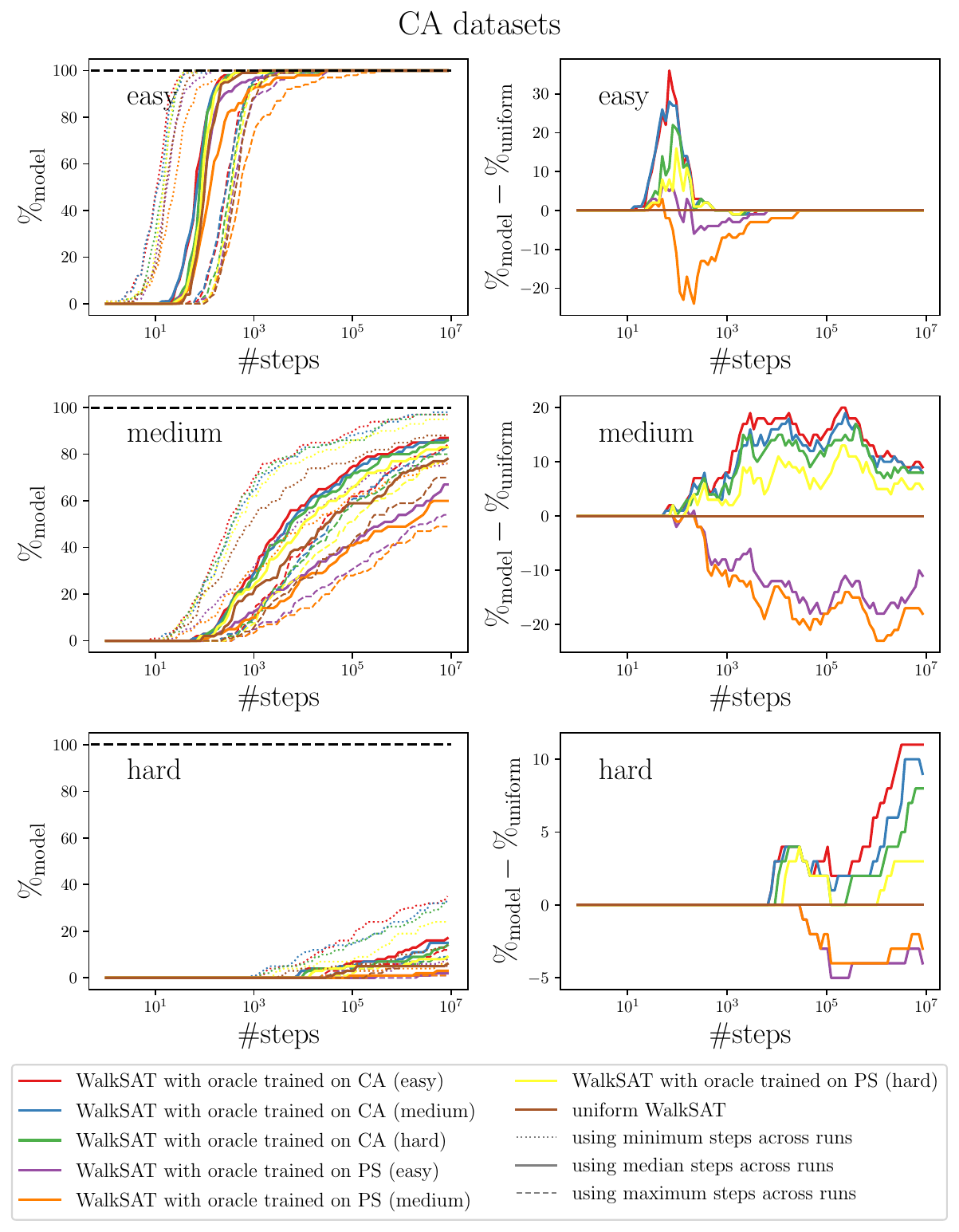}
    \hfill
    \includegraphics[width = 0.45 \columnwidth]{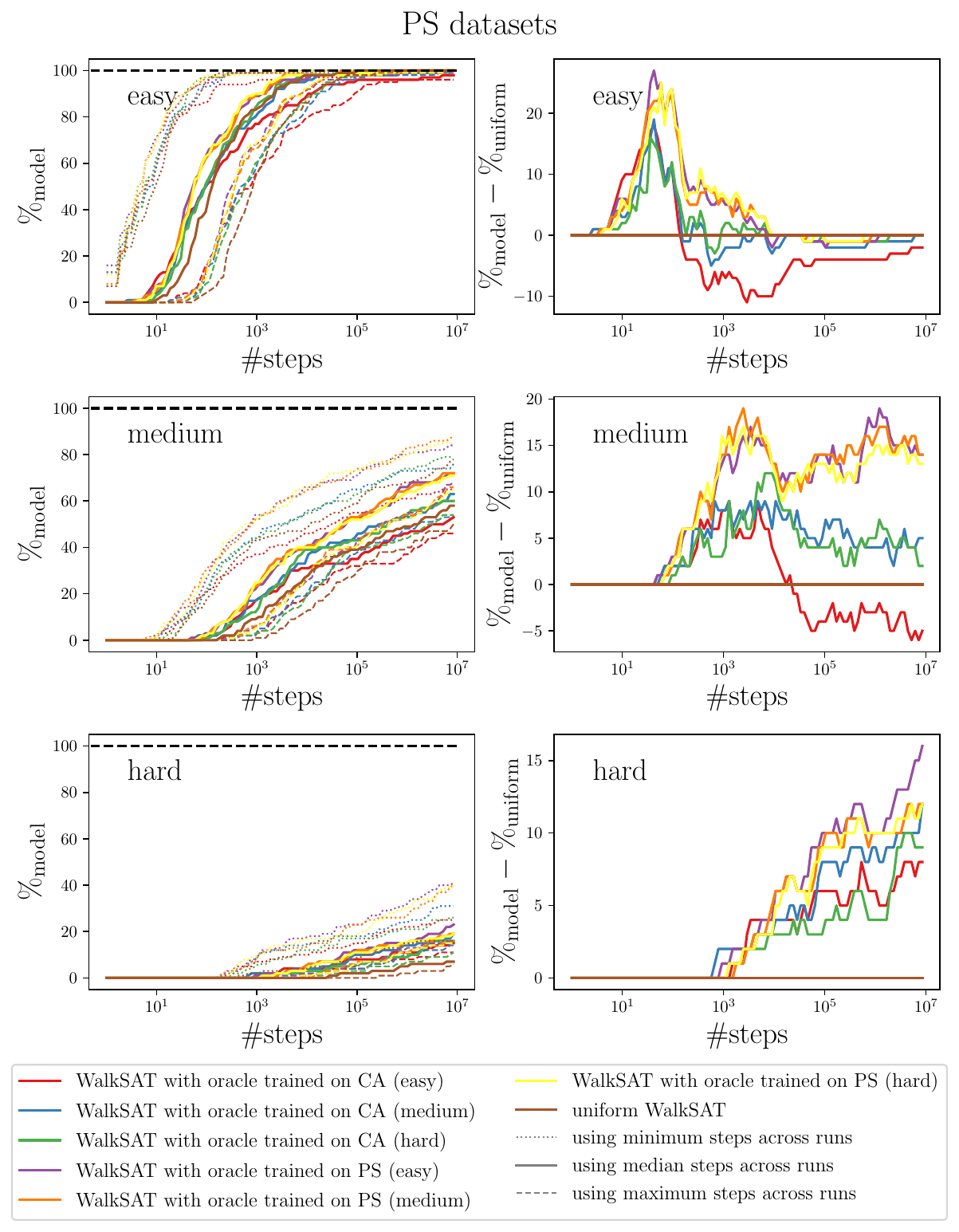}
    \caption{Cross-evaluations on the CA (left) and PS (right) datasets, with three distinct difficulty levels as indicated. The plots show the percentage of solved instances as a function of the number of steps taken for all trained models, alongside a uniform version serving as a baseline. Additionally, we plot the absolute difference in the percentage of solved instances between using the model and the uniform version.}
    \label{fig:industry_generalization}
\end{figure}

We conducted training on two pseudo-industrial dataset variants, each featuring three levels of difficulty. Cross-evaluations were performed on the CA and PS datasets, as shown in Table~\ref{tab:CA_cross} and Table~\ref{tab:PS_cross} respectively. This involved evaluating each dataset using all the acquired models, including those not specifically trained on the dataset. Particularly striking is the good generalization from easier to harder instances, i.e. training on easier instances than one evaluates on. It is surprising to see that in some cases, models trained on easier instances even outperform those trained specifically on harder ones. In Fig.~\ref{fig:industry_generalization} we plot the percentages of the solved instances as a function of the steps taken. In addition, we plot the relative improvement in this metric compared to the uniform version.

\begin{table}
  \caption{Cross-evaluation on the CA datasets with three different difficulty levels using different oracles. The best solver is highlighted for all metrics.}
  \label{tab:CA_cross}
  \centering
  \begin{tabular}{clccccc}
    \toprule
    Evaluated on & Used oracle & $\overline{\#}$ & $M(\#)$ & $\%_{\mathrm{M}}$ & $\%_{\mathrm{best}}$ & $\%_{\mathrm{worst}}$ \\
    \midrule
    \multirow{7}{*}{\rotatebox{90}{$\textrm{CA}_{\textrm{easy}}$}}
   &  $\textrm{CA}_{\textrm{easy}}$ & 1.83e2 & \textbf{7.88e1}       &         100      &       100     &       100 \\
    & $\textrm{CA}_{\textrm{medium}}$ & \textbf{1.67e2} & 8.30e1   &             100    &         100    &        100 \\
    &  $\textrm{CA}_{\textrm{hard}}$ & 1.75e2 & 9.42e1     &           100     &        100        &     100 \\
    &  $\textrm{PS}_{\textrm{easy}}$ & 3.39e2 & 1.17e2         &       100     &        100      &       100 \\
    & $\textrm{PS}_{\textrm{medium}}$ & 1.35e3 & 1.52e2   &             100     &        100    &         100 \\
    &  $\textrm{PS}_{\textrm{hard}}$ & 1.71e2 & 9.95e1      &          100       &      100         &    100 \\
    &uniform & 1.87e2 & 1.18e2  &              100       &      100     &        100 \\
    \midrule
    \multirow{7}{*}{\rotatebox{90}{$\textrm{CA}_{\textrm{medium}}$}}
    & $\textrm{CA}_{\textrm{easy}}$& \textbf{1.38e6} & \textbf{4.67e3} &            \textbf{87} &           97 &           \textbf{84} \\
    &$\textrm{CA}_{\textrm{medium}}$ & 1.48e6 & 5.94e3   &            86  &          \textbf{98}   &         83 \\
      & $\textrm{CA}_{\textrm{hard}}$ & 1.63e6 & 6.66e3    &           86    &        97      &      80 \\
    &  $\textrm{PS}_{\textrm{easy}}$ & 3.92e6 & 4.60e5      &        67       &     76     &       54 \\
    & $\textrm{PS}_{\textrm{medium}}$ & 4.36e6 & 1.65e6     &          60    &        76   &         49 \\
     & $\textrm{PS}_{\textrm{hard}}$ & 1.94e6 & 1.15e4  &              83   &         95   &         78 \\
    & uniform  & 2.52e6 & 3.52e4 &              78 &           88 &           70 \\
    \midrule
    \multirow{7}{*}{\rotatebox{90}{$\textrm{CA}_{\textrm{hard}}$}}
    & $\textrm{CA}_{\textrm{easy}}$ & \textbf{8.46e6} & 1.00e7 &\textbf{17} & \textbf{35} & \textbf{12} \\
     & $\textrm{CA}_{\textrm{medium}}$ & 8.68e6 & 1.00e7 & 15 & 33 &  9 \\
     & $\textrm{CA}_{\textrm{hard}}$ & 8.85e6 & 1.00e7   &            14  &          32       &     8\\
       & $\textrm{PS}_{\textrm{easy}}$ & 9.83e6 & 1.00e7       &        2    &        8       &     1 \\
        & $\textrm{PS}_{\textrm{medium}}$ & 9.76e6 & 1.00e7         &      3        &    6     &       1\\
         & $\textrm{PS}_{\textrm{hard}}$ & 9.17e6 &1.00e7    &           9      &      24    &        6\\
    & uniform & 9.45e6 & 1.00e7       &        6       &     14        &    5 \\
    \bottomrule
  \end{tabular}
\end{table}

\begin{table}
  \caption{Cross-evaluation on the PS datasets with three different difficulty levels using different oracles. The best solver is highlighted for all metrics.}
  \label{tab:PS_cross}
  \centering
  \begin{tabular}{clccccc}
    \toprule
    Evaluated on & Used oracle & $\overline{\#}$ & $M(\#)$ & $\%_{\mathrm{M}}$ & $\%_{\mathrm{best}}$ & $\%_{\mathrm{worst}}$\\

    \midrule
    \multirow{7}{*}{\rotatebox{90}{$\textrm{PS}_{\textrm{easy}}$}} 
    & $\textrm{CA}_{\textrm{easy}}$ & 2.73e5 & 1.17e2 & 98 & 99 & 96 \\
    & $\textrm{CA}_{\textrm{medium}}$ & 7.44e4 & 1.14e2 & 100 & 100 & 99 \\
    & $\textrm{CA}_{\textrm{hard}}$ & 3.38e4 & 1.10e2 & 100 & 100 & 99 \\
    & $\textrm{PS}_{\textrm{easy}}$ & 1.93e4 & \textbf{6.18e1} & 100 & 100 & 99 \\
    &$\textrm{PS}_{\textrm{medium}}$ & 1.46e4 & 6.80e1 & 100 & 100 & 100 \\
    & $\textrm{PS}_{\textrm{hard}}$ & 1.19e4 & 6.42e1 & 100 & 100 & 100 \\
    & uniform & \textbf{2.95e3} & 1.47e2 & 100 & 100 & 100 \\
    \midrule
    \multirow{7}{*}{\rotatebox{90}{$\textrm{PS}_{\textrm{medium}}$}}
    & $\textrm{CA}_{\textrm{easy}}$ & 5.00e6 & 4.85e6 & 53 & 65 & 46 \\
    & $\textrm{CA}_{\textrm{medium}}$ & 4.05e6 & 3.49e5 & 63 & 76 & 54 \\
    & $\textrm{CA}_{\textrm{hard}}$ & 4.14e6 & 4.95e5 & 60 & 79 & 54 \\
    & $\textrm{PS}_{\textrm{easy}}$ & \textbf{3.08e6} & 7.26e4 & \textbf{72} & 84 & \textbf{68} \\
    & $\textrm{PS}_{\textrm{medium}}$ & 3.09e6 & \textbf{6.08e4} & \textbf{72} & \textbf{88} & 66 \\
    & $\textrm{PS}_{\textrm{hard}}$ & 3.19e6 & 6.12e4 & 71 & 85 & 65 \\
    & uniform & 4.52e6 & 1.82e6 & 58 & 78 & 50 \\
    \midrule
    \multirow{7}{*}{\rotatebox{90}{$\textrm{PS}_{\textrm{hard}}$}}
     & $\textrm{CA}_{\textrm{easy}}$ & 8.67e6 & 1.00e7     &          15     &       25         &   11\\
      & $\textrm{CA}_{\textrm{medium}}$ & 8.38e6 &1.00e7      &         19       &     31      &     14\\
       & $\textrm{CA}_{\textrm{hard}}$ & 8.62e6 &1.00e7         &      16     &       26        &   11\\
        & $\textrm{PS}_{\textrm{easy}}$ & \textbf{7.99e6} &1.00e7          &     \textbf{23}    &      \textbf{41}        &   \textbf{16 }\\
        & $\textrm{PS}_{\textrm{medium}}$ & 8.18e6 & 1.00e7       &       19    &        40     &       \textbf{16}\\
        & $\textrm{PS}_{\textrm{hard}}$ & 8.20e6 & 1.00e7     &          19     &     40      &      15\\
    & uniform & 9.32e6 & 1.00e7 & 7 & 16 & 6 \\
    \bottomrule
  \end{tabular}
\end{table}

\end{document}